\documentclass[letterpaper, 10 pt, journal, twoside]{IEEEtran}

\usepackage{amssymb,amsfonts,amsmath,amscd,bbm} 
\usepackage{array}
\newcolumntype{?}{!{\vrule width 0.2mm}}
\usepackage{booktabs}
\usepackage{caption,subcaption} 
\usepackage{cite} 
\usepackage{enumitem} 
\usepackage[hidelinks]{hyperref} 
\usepackage{tikz} 
\usetikzlibrary{calc, arrows, shapes, positioning, decorations.pathreplacing, shadings, angles}
\pgfdeclareverticalshading{rainbow}{100bp}
{color(0bp)=(magenta);
 color(25bp)=(magenta);
 color(35bp)=(blue);
 color(45bp)=(cyan);
 color(55bp)=(green);
 color(65bp)=(yellow);
 color(75bp)=(red);
 color(100bp)=(red)}
\usepackage{acronym}

\newcommand{\benum}{\begin{enumerate}}
\newcommand{\eenum}{\end{enumerate}}

\newcommand{\mbf}{\mathbf}
\newcommand{\mbs}[1]{{\boldsymbol{#1}}}

\newcommand{\bbm}{\begin{bmatrix}}
\newcommand{\ebm}{\end{bmatrix}}

\newcommand{\beq}{\begin{equation}}
\newcommand{\eeq}{\end{equation}}
\newcommand{\beqn}{\begin{eqnarray}}
\newcommand{\eeqn}{\end{eqnarray}}

\newcommand{\Real}{\mathbb R}
\newcommand{\bcf}{\boldsymbol{\mathcal{F}}}
\renewcommand{\Vec}[1]{\underrightarrow{#1}}
\acrodef{FMCW}{Frequency-Modulated Continuous Wave}
\acrodef{GP}{Gaussian process}
\acrodef{ICP}{Iterative Closest Point}
\acrodef{IMU}{Inertial Measurement Unit}
\acrodef{PCA}{Principle Component Analysis}
\acrodef{WNOA}{White-Noise-on-Acceleration}

\title{Picking Up Speed: Continuous-Time Lidar-Only Odometry using Doppler Velocity Measurements}

\author{Yuchen Wu, David J. Yoon, Keenan Burnett, Soeren Kammel, Yi Chen, Heethesh Vhavle, and Timothy D. Barfoot
\thanks{This work was supported by the Natural Sciences and Engineering Research Council of Canada (NSERC).}%
\thanks{Yuchen Wu, David J. Yoon, Keenan Burnett, and Timothy D. Barfoot are with the University of Toronto Institute for Aerospace Studies (UTIAS), University of Toronto, Toronto, ON M3H5T6, Canada
  {\tt\footnotesize yuchen.wu; david.yoon; keenan.burnett; tim.barfoot [@robotics.utias.utoronto.ca]}}%
\thanks{Soeren Kammel, Yi Chen, and Heethesh Vhavle are with Aeva Inc., Mountain View, CA 94043, USA
  {\tt\footnotesize soeren; ychen; heethesh [@aeva.ai]}}%
}

\begin{document}

\maketitle
\bibliographystyle{IEEEtran}
\thispagestyle{empty}
\pagestyle{empty}

\begin{abstract}
  Frequency-Modulated Continuous-Wave (FMCW) lidar is a recently emerging technology that additionally enables per-return instantaneous relative radial velocity measurements via the Doppler effect. In this letter, we present the first continuous-time lidar-only odometry algorithm using these Doppler velocity measurements from an FMCW lidar to aid odometry in geometrically degenerate environments. We apply an existing continuous-time framework that efficiently estimates the vehicle trajectory using Gaussian process regression to compensate for motion distortion due to the scanning-while-moving nature of any mechanically actuated lidar (FMCW and non-FMCW). We evaluate our proposed algorithm on several real-world datasets, including publicly available ones and datasets we collected. Our algorithm outperforms the only existing method that also uses Doppler velocity measurements, and we study difficult conditions where including this extra information greatly improves performance. We additionally demonstrate state-of-the-art performance of lidar-only odometry with and without using Doppler velocity measurements in nominal conditions. Code for this project can be found at: \url{https://github.com/utiasASRL/steam_icp}.
\end{abstract}


\IEEEpeerreviewmaketitle

\section{Introduction}

Multi-beam lidars have become a common addition to the sensor suite of an autonomous vehicle. Estimation algorithms to handle the long-range 3D measurements (i.e., point clouds) produced by these sensors have also matured, and are capable of producing highly accurate motion estimates, often at a sub-decimeter level of accuracy for localization \cite{burnett_ral22}.

Lidar motion estimation performs exceptionally when there exists sufficient geometric structure in the surroundings to uniquely constrain all six degrees of freedom of the vehicle pose. In contrast, even the best estimators will struggle and even fail in geometrically degenerate environments. Long tunnels, highways with a barren landscape, and bridges are typical examples of extreme conditions where prior knowledge of the vehicle kinematics is insufficient to compensate for the lack of geometric information. A common solution in such situations is to rely on an additional sensor such as an \ac{IMU} \cite{zhang_ar17, ye_icra19}.

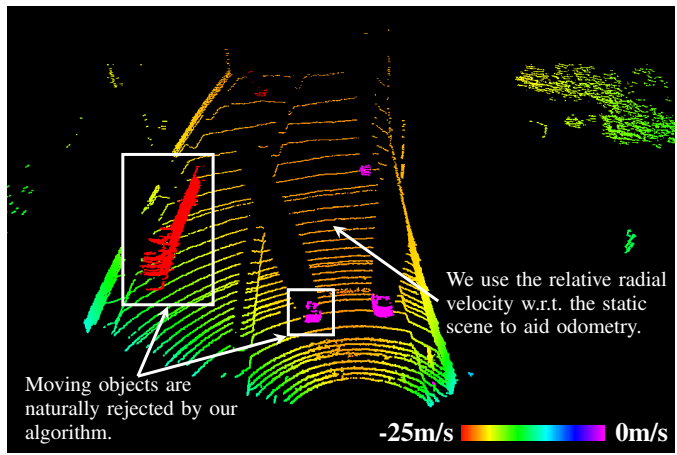
\begin{figure}[t]
  \begin{tikzpicture}[
      boundary/.style={>=latex,blue, line width=1.25pt},
      custom arrow/.style={single arrow, minimum height=12mm, minimum width=1mm, single arrow head extend=1.5mm, scale=0.15, fill=blue},
      block/.style={rectangle, white, minimum width=4mm, align=center, inner sep=0pt},
      rotate border/.style={shape border uses incircle, shape border rotate=#1},
    ]

    \node[inner sep=0pt] (p2) at (0mm, 0mm) {\includegraphics[trim=800 120 700 150, clip, width=\columnwidth]{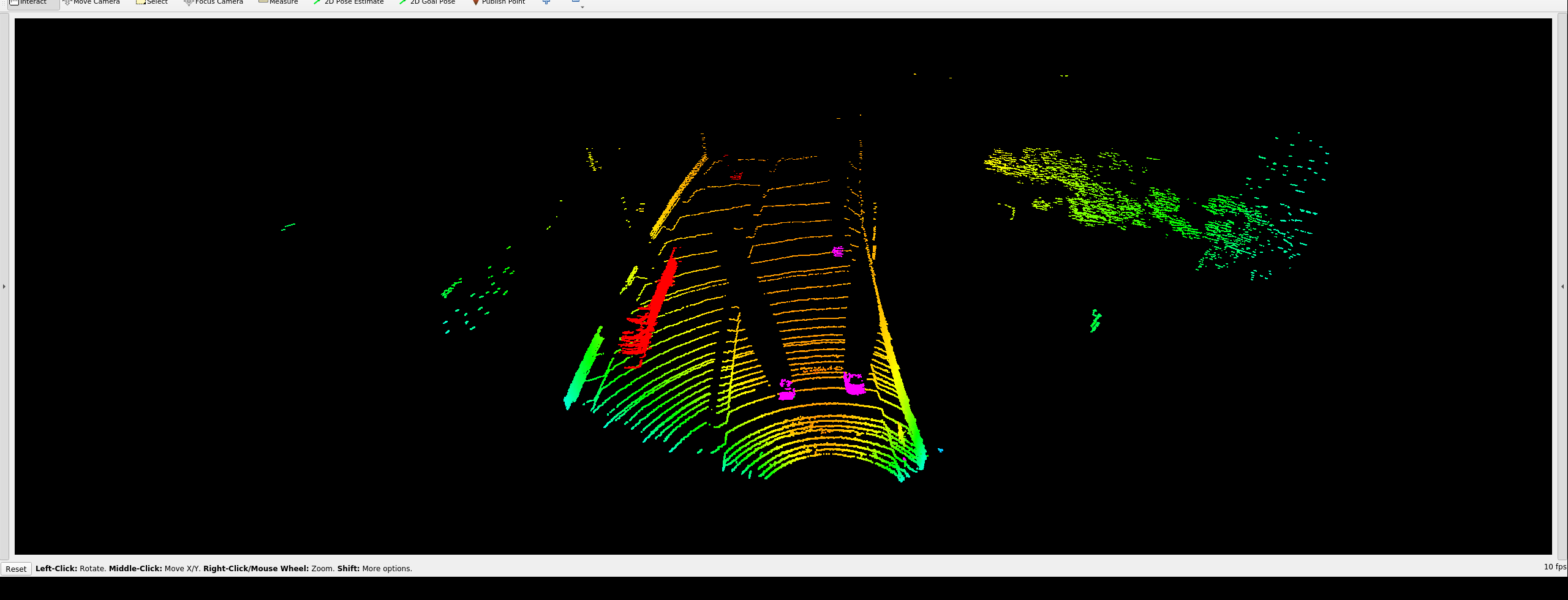}};

    \node[block] (l1) at (10mm, -27mm) {\textbf{-25m/s}};
    \node[block] (l2) at ($(l1.center) + (30mm, 0mm)$) {\textbf{0m/s}};
    \shade[shading=rainbow,shading angle=90] ($(l1.center) + (6mm, -1mm)$) rectangle ($(l2.center) + (-5mm, 1mm)$);

    \node[block, white, text width=30mm, align=left, font=\footnotesize] (t2) at (29mm, -10mm) {We use the relative radial velocity w.r.t. the static scene to aid odometry.};
    \draw [-stealth, white, line width=0.4mm] ($(t2.north west) + (-1mm, -4mm)$) to ++(-14mm, 10mm) {};

    \node[block, white, text width=30mm, align=left, font=\footnotesize] (t1) at (-27mm, -24mm) {Moving objects are naturally rejected by our algorithm.};
    \node[block, white, draw, line width=0.4mm, minimum height=20mm, minimum width=12mm] (d1) at (-23mm, 0mm) {};
    \node[block, white, draw, line width=0.4mm, minimum height=6mm, minimum width=6mm] (d2) at (-4mm, -11mm) {};
    \draw[-stealth, white, line width=0.4mm] ($(t1.north) + (0mm, 0mm)$) to ($(d1.south) + (0mm, 0mm)$) {};
    \draw[-stealth, white, line width=0.4mm] ($(t1.north) + (0mm, 0mm)$) to ($(d2.south west) + (0mm, 0mm)$) {};

  \end{tikzpicture}
  \caption{An example lidar frame from the Aeva Aeries I FMCW Lidar colored by the measured relative radial velocity (i.e., Doppler velocity). Vehicles driving in the same direction are colored in magenta as they have the same velocity as the ego-vehicle, while vehicles driving in the opposite direction are in red. They can be clearly distinguished from the underlying static scene. Our proposed algorithm uses the relative radial velocity with respect to the static scene to aid odometry. Moving objects are naturally rejected by our algorithm using robust estimation techniques.}
  \label{fig:lidar_data_example}
\end{figure}

\ac{FMCW} lidar is a recently emerging technology \cite{behroozpour_cm17, royo_as19} that provides a promising alternative solution for geometrically degenerate environments. While capable of producing dense point clouds comparable to a typical time-of-flight lidar, \ac{FMCW} lidars also measure the relative velocity between each measured point along the radial direction via the Doppler effect, which we call a \textit{Doppler velocity measurement}.
Figure~\ref{fig:lidar_data_example} depicts an example lidar frame\footnote{Throughout this letter, we refer to the aggregation of points over one full lidar field-of-view as a lidar frame.} colored by the Doppler velocity of each point. Hexsel et al. \cite{hexsel_rss22} recently showed that these Doppler velocity measurements are beneficial for lidar odometry in an \ac{ICP}-based algorithm.


In this letter, we improve upon the existing work \cite{hexsel_rss22} by incorporating the Doppler velocity measurements in a continuous-time estimation framework. Continuous-time estimation allows for each measurement to be associated with its actual time of acquisition, avoiding the need for an \ac{IMU} to correct the motion distortion of a lidar frame due to the scanning-while-moving nature of mechanically actuated lidars.
Similar to \cite{hexsel_rss22}, we present a Doppler velocity factor that can be applied in conjunction with the usual point-to-plane factor for frame-to-map alignment. However, our proposed factor differs from theirs in that it is applied to the vehicle's body-centric velocity as opposed to pose since body-centric velocity is also part of our estimated state. We evaluate our lidar odometry algorithm on several real-world datasets, including publicly available ones collected using a non-FMCW lidar and datasets we collected using an FMCW lidar. Through our evaluation, we demonstrate overall state-of-the-art lidar-only odometry performance with and without using Doppler velocity measurements under both nominal and geometrically degenerate conditions.

\section{Related Work}

Lidar motion estimation typically adopts a point cloud registration approach using a variant of ICP \cite{besl_pami92, pomerleau_15}. Most lidar motion estimation pipelines can be divided into a data processing front-end and a state estimation back-end \cite{cadena_tro16}.

The front-end processes the raw lidar frames, which includes keypoint/feature extraction, global/local map building, and data association. LOAM \cite{zhang_ar17} extracts edge and plane features and matches them via nearest-neighbor association. SuMa \cite{behley_rss18} matches raw frames to a surfel map using projective data association. Recently, Vizzo et al. \cite{vizzo_icra21} introduced a triangle mesh map representation and a ray-casting-based point-to-mesh association approach. Other alternatives have been proposed in \cite{shan_iros18, chen_iros19, kovalenko_ecmr19, lin_icra20, chen_ral20, zheng_ral21, pan_icra21}, though we do not discuss them in detail in the interest of space.


The back-end leverages the processed data to estimate the vehicle state over time, seen as the vehicle trajectory. Existing approaches are different in how they formulate and estimate the trajectory. Discrete-time estimators formulate a trajectory where there is a temporal state (i.e., marginal) that corresponds to the acquisition time of every measurement. However, typical multi-beam lidars are mechanically actuated and produce thousands of measurements for every lidar frame. Consequently, each measurement may have a unique timestamp. Even after preprocessing into sparser keypoints, it is often not feasible to have a discrete state estimate at each measurement time. One option is to correct for the motion of each frame using an additional sensor such as an \ac{IMU} \cite{zhang_ar17, ye_icra19}.



Alternatively, the vehicle trajectory can be estimated as a continuous function of time. The most straightforward approach is to apply linear interpolation between discrete states \cite{bosse_icra09, zhang_ar17, park_icra18}. However, using linear interpolation between poses restricts the trajectory to have a piece-wise constant velocity, which cannot accurately represent trajectories undergoing high-frequency changes in velocity. Dellenbach et al. \cite{dellenbach_icra22} address this limitation by allowing the vehicle poses to be discontinuous between lidar frames, trading trajectory smoothness for a closer representation. Another option for higher representational power without losing smoothness is to represent the trajectory using temporal basis functions \cite{furgale_icra12,anderson_icra13}. Referred to as a parametric methods, the trajectory is parameterized by the associated basis-function weights and has been demonstrated several times for lidar motion estimation using B-splines \cite{zlot_jfr14, kaul_jfr16, alismail_icra14, droeschel_icra18}.
Nonparametric methods instead estimate the continuous-time trajectory as a \ac{GP} \cite{tong_ijrr13}. Barfoot et al. \cite{barfoot_rss14} demonstrated that through careful selection of the underlying GP prior and a Markovian state, we can benefit from the usual sparsity exploited in discrete-time estimation. Anderson and Barfoot \cite{anderson_iros15} extended this idea for trajectory estimation in $SE(3)$. A desirable feature of their approach is that the \ac{GP} prior is made to be physically motivated (e.g., white noise on acceleration). Lidar motion estimation has been demonstrated several times using various GP priors \cite{tang_crv18,tang_ral19,wong_ral20,burnett_ral22}.

All works discussed thus far consider lidars that only produce point clouds (optionally with intensity information). Advancements in \ac{FMCW} technology have enabled a new type of lidar, i.e., the Aeva Aeries I \ac{FMCW} Lidar \cite{aeva}, that can additionally measure the relative radial velocity (Doppler velocity) of each point. Hexsel et al. \cite{hexsel_rss22} presented an ICP-based algorithm that uses the Doppler velocity measurements and demonstrated improved performance in environments with insufficient geometric structure.

The Aeva Aeries I FMCW Lidar is mechanically actuated, similar to existing multi-beam lidars. Hexsel et al. \cite{hexsel_rss22} applied a discrete-time estimator, which requires the lidar frames to be corrected for motion beforehand using an \ac{IMU}. We improve upon their work by formulating the back-end estimator using GP regression in $SE(3)$ \cite{anderson_iros15}. Although the GP regression approach has been applied to lidar odometry before \cite{tang_crv18,tang_ral19,wong_ral20,burnett_ral22}, the benefits of including Doppler velocity measurements have not been demonstrated. Compared to the work of Hexsel et al. \cite{hexsel_rss22}, we demonstrate state-of-the-art performance in lidar odometry without using additional sensors.

Furthermore, the \ac{FMCW} technology can also be applied to radars, and there has been some work using an \ac{FMCW} radar for motion estimation \cite{kellner_itsc13, vivet_sens13}. Both Kellner et al. \cite{kellner_itsc13} and Vivet et al. \cite{vivet_sens13} use the Doppler velocity measurements from an \ac{FMCW} radar to estimate the velocity of the ego-vehicle, which is the same as what they are used for in this work.

\section{Methodology}

Our lidar odometry algorithm adopts the conventional \ac{ICP}-based frame-to-map approach while incorporating Doppler velocity measurements in a sliding-window implementation.

\subsection{Data Processing Front-End}
\label{sec:front-end}



We follow the approach of Dellenbach et al. \cite{dellenbach_icra22}. Keypoints are extracted from each lidar frame via voxel grid downsampling. We use a grid size of 1.5m and keep one random point in each voxel. Our local map is a point cloud accumulated from the most recent frames and cropped to be within 100m of the latest estimate of the vehicle after each frame update. The local map is stored in a sparse voxel grid structure with a 1m grid size and a maximum of 20 points per voxel. We use point-to-plane ICP for frame-to-map matching. Each frame point is associated with a map point via nearest-neighbor association, and the corresponding plane normal is computed by applying \ac{PCA} to the 20 closest neighbors of the associated map point.



\subsection{Trajectory Estimation Back-End}
\label{sec:trajectory-estimation}

\begin{figure}[htbp]
  \centering
  \begin{tikzpicture}[
      line cap=round,
      graphedge/.style={>=latex, line width=1pt},
      state/.style={draw, thick, line width=1pt, isosceles triangle,isosceles triangle apex angle=45, minimum size=4mm, inner sep=0pt, outer sep=0pt},
      interpolated/.style={state, minimum size=2mm},
      dot/.style={draw, inner sep=0, circle, fill, minimum size=2mm, line width=0pt},
    ]

    \node[] (p1) {};
    \node[] (p2) [right=10mm of p1.center] {};
    \node[] (p3) [right=10mm of p2.center] {};
    \node[] (p4) [right=10mm of p3.center] {};
    \node[] (p5) [right=10mm of p4.center] {};
    \node[] (p6) [right=10mm of p5.center] {};
    \node[] (p7) [right=10mm of p6.center] {};
    \node[] (p8) [right=10mm of p7.center] {};
    \draw[graphedge, dotted] (p1) -- (p2);
    \begin{scope}
      \clip (p2) circle (6mm);
      \draw[graphedge] (p1) -- (p2);
    \end{scope}
    \draw[graphedge] (p2) edge node[dot, midway, red, xshift=1mm] {} (p3);
    \draw[graphedge] (p3) edge node[dot, midway, red] {} (p4);
    \draw[graphedge] (p4) edge node[dot, midway, red] {} (p5);
    \draw[graphedge] (p5) edge node[dot, midway, red, xshift=1mm] {} (p6);
    \draw[graphedge] (p6) edge node[dot, midway, red] {} (p7);
    \draw[graphedge] (p7) edge node[dot, midway, red] {} (p8);

    \draw[graphedge, yshift=-3mm] (p3.center) -- ($(p3.center) + (0, 7mm)$);
    \draw[graphedge, yshift=-3mm] (p4.center) -- ($(p4.center) + (0, 7mm)$);
    \draw[graphedge, yshift=-3mm] (p5.center) -- ($(p5.center) + (0, 7mm)$);
    \draw[graphedge, yshift=-3mm] (p6.center) -- ($(p6.center) + (0, 7mm)$);
    \draw[graphedge, yshift=-3mm] (p7.center) -- ($(p7.center) + (0, 7mm)$);
    \draw[graphedge, yshift=-3mm] (p8.center) -- ($(p8.center) + (0, 7mm)$);
    \node[dot, green] at ($(p3.center) + (0, 7mm)$) {};
    \node[dot, green] at ($(p4.center) + (0, 7mm)$) {};
    \node[dot, green] at ($(p5.center) + (0, 7mm)$) {};
    \node[dot, green] at ($(p6.center) + (0, 7mm)$) {};
    \node[dot, green] at ($(p7.center) + (0, 7mm)$) {};
    \node[dot, green] at ($(p8.center) + (0, 7mm)$) {};

    \node[] (m1) at ($(p2.center) + (1mm, -11mm)$) {};
    \node[] (m2) at ($(p2.center) + (4mm, -13mm)$) {};
    \node[] (m3) at ($(p2.center) + (6mm, -15mm)$) {};
    \node[] (m4) at ($(p2.center) + (8.5mm, -17mm)$) {};
    \node[] (m5) at ($(p2.center) + (8mm, -19mm)$) {};
    \node[] (m6) at ($(p2.center) + (6mm, -22mm)$) {};
    \draw[graphedge] (p3) edge[bend left=20] node[dot, midway, blue] {} (m1);
    \draw[graphedge] (p4) edge[bend left=20] node[dot, midway, blue] {} (m2);
    \draw[graphedge] (p5) edge[bend left=20] node[dot, midway, blue] {} (m3);
    \draw[graphedge] (p6) edge[bend left=20] node[dot, midway, blue] {} (m4);
    \draw[graphedge] (p7) edge[bend left=20] node[dot, midway, blue] {} (m5);
    \draw[graphedge] (p8) edge[bend left=20] node[dot, midway, blue] {} (m6);

    %
    \node[dot, black, minimum size=1mm] (mm1) at ($(p2.center) + (1mm, -11mm)$) {};
    \node[dot, black, minimum size=1mm] (mm2) at ($(p2.center) + (4mm, -13mm)$) {};
    \node[dot, black, minimum size=1mm] (mm3) at ($(p2.center) + (6mm, -15mm)$) {};
    \node[dot, black, minimum size=1mm] (mm4) at ($(p2.center) + (8mm, -17mm)$) {};
    \node[dot, black, minimum size=1mm] (mm5) at ($(p2.center) + (8mm, -19mm)$) {};
    \node[dot, black, minimum size=1mm] (mm6) at ($(p2.center) + (6mm, -22mm)$) {};
    \node[dot, black, minimum size=1mm] (mm11) at ($(p2.center) + (-2mm, -11mm)$) {};
    \node[dot, black, minimum size=1mm] (mm12) at ($(p2.center) + (0mm, -13mm)$) {};
    \node[dot, black, minimum size=1mm] (mm13) at ($(p2.center) + (3mm, -16mm)$) {};
    \node[dot, black, minimum size=1mm] (mm14) at ($(p2.center) + (5mm, -19mm)$) {};
    \node[dot, black, minimum size=1mm] (mm15) at ($(p2.center) + (2mm, -20mm)$) {};
    \node[dot, black, minimum size=1mm] (mm16) at ($(p2.center) + (-1mm, -18mm)$) {};

    \draw [decorate, decoration={brace,amplitude=10pt}] ($(p2.center) + (2mm, 12mm)$) -- ($(p5.center) + (2mm, 12mm)$) node [black,midway,yshift=6mm] {$(k-1)^{\text{th}}$ Frame};
    \draw [decorate, decoration={brace,amplitude=10pt}] ($(p5.center) + (2mm, 12mm)$) -- ($(p8.center) + (2mm, 12mm)$) node [black,midway,yshift=6mm] {$k^{\text{th}}$ Frame};

    \node[state, fill=gray, label=above:$\mbf{x}_{k-2}$] (s0) at (p2.center) {};
    \node[interpolated, fill=white] at (p3.center) {};
    \node[interpolated, fill=white] at (p4.center) {};
    \node[state, fill=white, label=above right:$\mbf{x}_{k-1}$] (s1) at (p5.center) {};
    \node[interpolated, fill=white] at (p6.center) {};
    \node[interpolated, fill=white] at (p7.center) {};
    \node[state, fill=white, label=above right:$\mbf{x}_{k}$] (s2) at (p8.center) {};

    \coordinate (legend) at ($(p1.west)!0.5!(p8.east) + (0mm, -30mm)$);
    \node[state, left=35mm of legend, label={[font=\footnotesize, label distance=0.5mm]right:Estimated State}] (c2) {};
    \node[interpolated, left=8mm of legend, yshift=0mm, label={[font=\footnotesize, label distance=0.5mm]right:Interpolated State}] {};
    \node[dot, black, minimum size=1mm, left=-19.5mm of legend, yshift=0mm, label={[font=\footnotesize, label distance=1mm]right:Local Map Point}] {};

    \node[dot, red, left=35mm of legend, yshift=-5mm, label={[font=\footnotesize, label distance=0.5mm]right:Motion Prior $\phi_{\text{mp}}$}] {};
    \node[dot, blue, left=8mm of legend, yshift=-5mm, label={[font=\footnotesize, label distance=0.5mm]right:Point-to-Plane $\phi_{\text{p2p}}$}] {};
    \node[dot, green, left=-20mm of legend, yshift=-5mm, label={[font=\footnotesize, label distance=0.5mm]right:Doppler Velocity $\phi_{\text{dv}}$}] {};
  \end{tikzpicture}
  \caption{This diagram shows the states and factors involved in our sliding-window estimation with an example window size of two. We query the trajectory states at the acquisition time of each keypoint in each lidar frame. The motion prior factor $\phi_{\text{mp}}$ connects neighboring trajectory states. The point-to-plane factor $\phi_{\text{p2p}}$ requires local map information while the Doppler velocity factor $\phi_{\text{dv}}$ does not. For computational efficiency, we implement this sliding-window estimation by introducing one estimated state at the end of each lidar frame and using \ac{GP} interpolation to obtain states at other measurement times. This way, motion prior factors between interpolated and estimated/interpolated states are merged into factors between estimated states. Point-to-plane and Doppler velocity factors applied to interpolated states depend on the adjacent estimated states. Past states falling outside the window are marginalized and no longer updated.}
  \label{fig:cticp_factor_graph}
\end{figure}
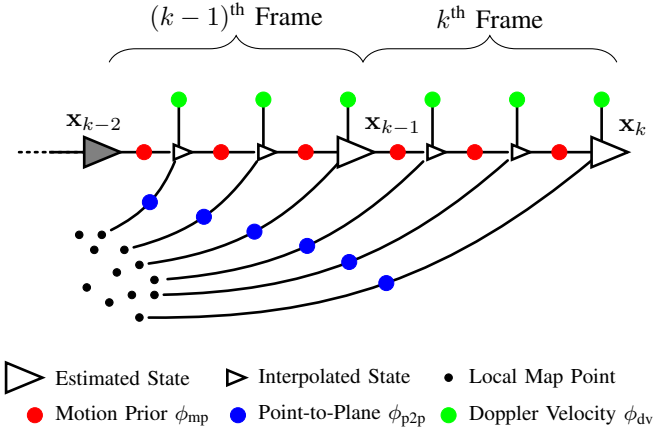

\textbf{Notation:} We denote $\Vec{\bcf}_i$ to be the inertial reference frame, $\Vec{\bcf}_v$ to be the vehicle reference frame, and $\Vec{\bcf}_\ell$ to be the lidar reference frame. Let $\mbf{T}_{vi} \in SE(3)$ be the inertial-to-vehicle transformation (i.e., the vehicle pose) and $\mbs{\varpi}^{iv}_v = \begin{bmatrix} {\mbs{\nu}^{iv}_v}^T & {\mbs{\omega}^{iv}_v}^T \end{bmatrix}^T \in \mathbb{R}^6$ be the vehicle body-centric velocity where $\mbs{\nu}^{iv}_v$ and $\mbs{\omega}^{iv}_v$ are translational and rotational velocities, respectively\footnote{The superscripts and subscripts follow the convention in \cite{barfoot_se17}.}. Furthermore, let $\mbf{T}_{\ell v}$ be the fixed vehicle-to-lidar transformation assuming the lidar is rigidly mounted on the vehicle, $\mbf{q}$ be a homogeneous point from a lidar frame (expressed in $\Vec{\bcf}_\ell$), and $\mbf{p}$ be a homogeneous point in the local map (expressed in $\Vec{\bcf}_i$).

We follow previous work \cite{anderson_iros15} to represent a continuous-time trajectory as a \acf{GP}. Our trajectory is $\mbf{x}(t) = \{\mbf{T}_{vi}(t), \mbs{\varpi}^{iv}_v(t) \}$. We use the \ac{WNOA} motion prior
\begin{equation}
  \begin{split}
    \dot{\mathbf{T}}_{vi}(t) &= \mbs{\varpi}^{iv}_v(t)^\wedge \mathbf{T}_{vi}(t), \\
    \dot{\mbs{\varpi}}^{iv}_v(t) &= \mathbf{w}(t),\quad \mathbf{w}(t) \sim \mathcal{GP}(\mathbf{0}, \mathbf{Q}_c\delta(t-\tau)),
  \end{split}
  \label{eq:se3_prior}
\end{equation}
where $\mbf{w}(t) \in \Real^6$ is a zero-mean, white-noise GP. The prior is applied in a piecewise fashion across an underlying discrete trajectory of pose-velocity state pairs, $\mbf{x}_k$, each corresponding to the end time of the $k^{\text{th}}$ lidar frame\footnote{$\mbf{x}_0$ is at the start of the first frame.}. Note that the period of the $k^{\text{th}}$ frame lies between $\mbf{x}_{k-1}$ and $\mbf{x}_k$. For a keypoint $\mbf{q}_j$ with acquisition time $t_j$ where $t_{k-1} < t_j < t_k$, we associate it to an interpolated state $\mbf{x}(t_j)$ that depends on $\mbf{x}_{k-1}$ and $\mbf{x}_k$ through \ac{GP} interpolation. Interpolation is done efficiently through our choice of motion prior and Markovian state \cite{anderson_iros15}.

We jointly align the most recent five frames to the local map in a sliding window by optimizing for the states $\mbf{x}_{k-5}$ to $\mbf{x}_{k}$. Figure~\ref{fig:cticp_factor_graph} illustrates the states and factors in our problem. We refer readers to Anderson and Barfoot \cite{anderson_iros15} for details on the motion prior factor and discuss the measurement factors below.

\subsubsection{Point-to-Plane Factor}

The point-to-plane factor is
\begin{eqnarray}
  \label{eq:p2p}
  \phi_{\text{p2p}} &=& \rho \left( \sqrt{ \alpha^4 \; e^2_{\text{p2p}}} \right), \\
  e_{\text{p2p}} &=& \mbf{n}^T \mbf{D} \left( \mbf{p} - \mbf{T}_{vi}(t)^{-1} \mbf{T}_{\ell v}^{-1} \mbf{q} \right),
\end{eqnarray}
with $\mbf{q}$ measured at time $t$ and $\mbf{T}_{vi}(t)$ being the vehicle pose queried from the trajectory. $\mbf{p}$ is the nearest neighbor of $\mbf{q}$ in the local map, $\mbf{n}$ is the surface normal of $\mbf{p}$, and $\alpha = (\sigma_2 - \sigma_3) / \sigma_1$ \cite{dellenbach_icra22, deschaud_icra18} is a heuristic weight to favour planar neighborhoods. $\mbf{D}$ is a constant projection that removes the fourth homogeneous element. $\rho(\cdot)$ is a robust cost function chosen to be Cauchy with $c=0.5$ \cite{mactavish_crv15}, and we discard any measurement resulting in $e_{\text{p2p}} \geq 0.5\text{m}$.

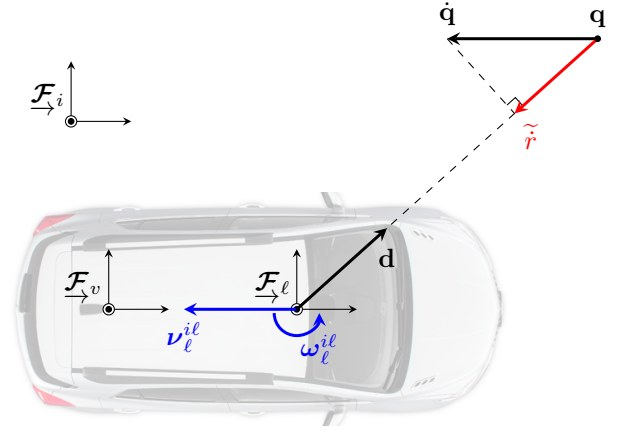
\begin{figure}[t]
  \begin{tikzpicture}[
      custom arrow/.style={-stealth, line width=0.4mm},
      frame/.style={inner sep=0pt,minimum size=16mm, path picture={
              \draw[-stealth] (0,0) -- (8mm, 0mm);
              \draw[-stealth] (0,0) -- (0mm, 8mm);
              \node[circle, draw, inner sep=0pt, minimum size=1.5mm, fill=white] at (0,0) {};
              \node[circle, draw, inner sep=0pt, minimum size=0.75mm, fill=black] at (0,0) {};
            }},
      block/.style={rectangle, white, minimum width=4mm, align=center, inner sep=0pt},
      rotate border/.style={shape border uses incircle, shape border rotate=#1},
    ]

    \coordinate (inertial) at (5mm,25mm);
    \coordinate (vehicle) at (10mm,0);
    \coordinate (sensor) at (35mm,0);
    \coordinate (point) at (75mm,36mm);
    \coordinate (point velocity) at ($(point) + (-20mm, 0mm)$);
    \coordinate (radial) at ($(point)!(point velocity)!(sensor)$);

    \node[opacity=0.3, xshift=20mm] at (vehicle) {\includegraphics[trim=0 100 0 100, clip, width=0.8\columnwidth]{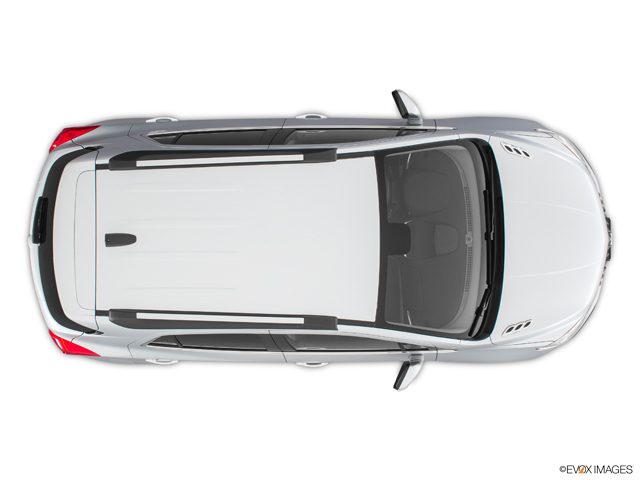}};

    \node[frame, label={[label distance=-12mm]north west:$\Vec{\bcf}_{i}$}] at (inertial) {};
    \node[frame, label={[label distance=-12mm]north west:$\Vec{\bcf}_{v}$}] at (vehicle) {};
    \node[frame, label={[label distance=-12mm]north west:$\Vec{\bcf}_{\ell}$}] at (sensor) {};

    \draw[custom arrow, blue] (sensor) to ++(-15mm, 0mm) node[label={[label distance=-1mm]south:$\mbs{\nu}_{\ell}^{i\ell}$}] {};
    \draw[custom arrow, blue] ($(sensor) + (-3mm, -0.5mm)$) arc[start angle=180, end angle=360, x radius=3mm, y radius=3mm] node[label={[label distance=0mm]south:$\mbs{\omega}_{\ell}^{i\ell}$}] {};

    \draw[dashed] (sensor) to (point) node[label={[label distance=-1mm]north:$\mbf{q}$}] {};
    \draw[dashed] (point velocity) to (radial) node {};
    \draw (radial) pic [black, draw, angle radius=1.5mm] {right angle = point velocity--radial--point};

    \draw[custom arrow] (point) to (point velocity) node[label={[label distance=-1mm]north:$\dot{\mbf{q}}$}] {};
    \draw[custom arrow, red] (point) to (radial) node[label={[label distance=-2mm]south east:$\widetilde{\dot{r}}$}] {};
    \draw[custom arrow] (sensor) to ($(sensor)!0.3!(point)$) node[label={[label distance=0mm]south:$\mbf{d}$}] {};

    \node[circle, draw, inner sep=0pt, minimum size=0.75mm, fill=black] at (point) {};

  \end{tikzpicture}
  \caption{This diagram provides a graphical illustration of the Doppler velocity error term derivation. $\protect\Vec{\bcf}_{i}$, $\protect\Vec{\bcf}_{v}$, and $\protect\Vec{\bcf}_{\ell}$ are the inertial, vehicle, and lidar reference frames, respectively. $\mbf{q}$ and $\dot{\mbf{q}}$ are the measured point position and velocity expressed in $\protect\Vec{\bcf}_{\ell}$, respectively. $\mbf{d}$ is a unit vector along the direction of $\mbf{q}$, and $\widetilde{\dot{r}}$ is the relative radial velocity according to $\mbs{\varpi}^{i\ell}_\ell = \begin{bmatrix} {\mbs{\nu}^{i\ell}_\ell}^T & {\mbs{\omega}^{i\ell}_\ell}^T \end{bmatrix}^T$. See Section~\ref{sec:trajectory-estimation} for the expression of $\mbf{d}$ and $\widetilde{\dot{r}}$.}
  \label{fig:dv_derivation}
\end{figure}

\begin{figure*}[ht]
  \centering
  \begin{subfigure}{0.24\textwidth}
    \includegraphics[trim=315 0 315 0, clip, width=\columnwidth]{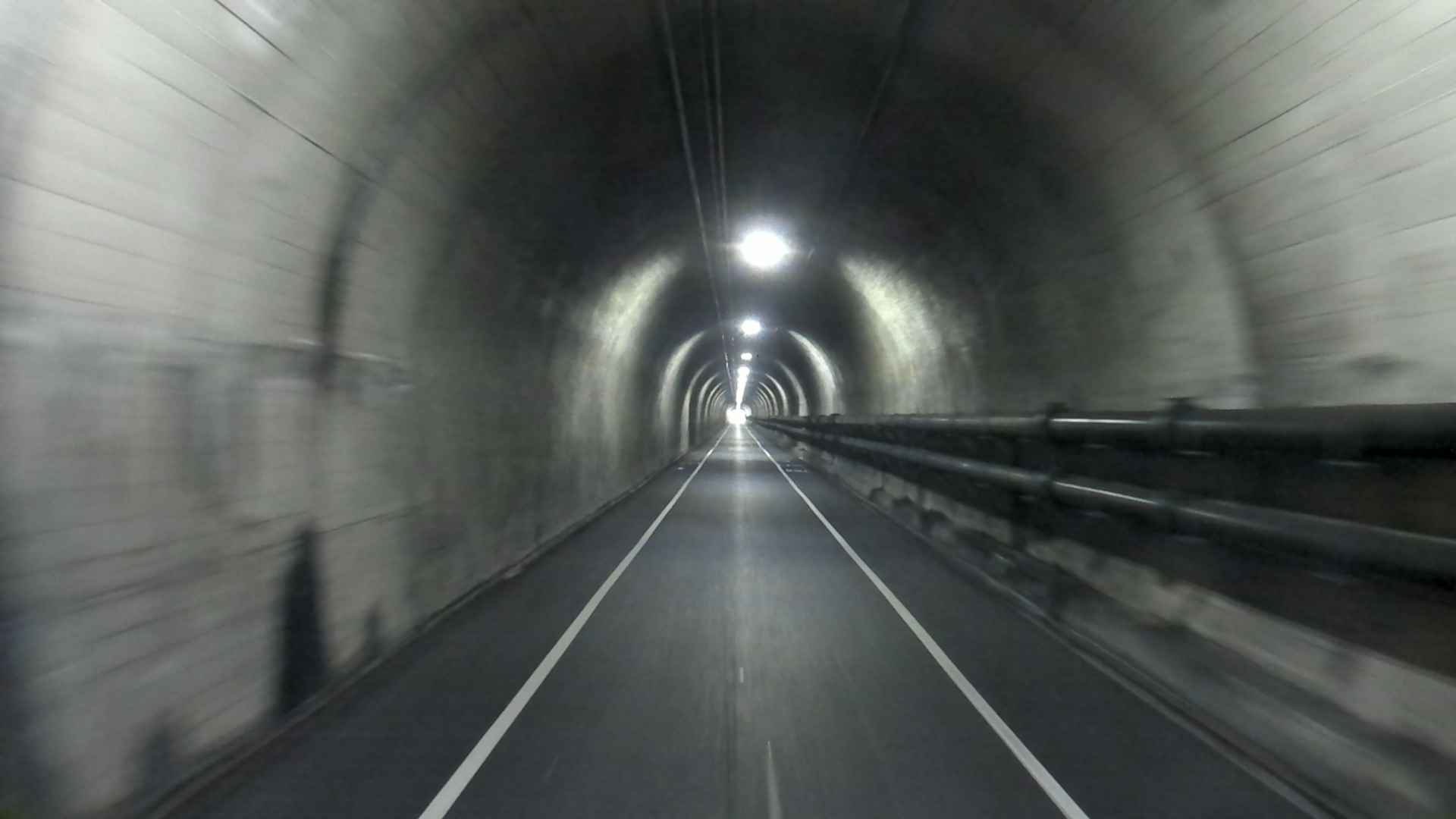}
    \caption{Baker-Barry Tunnel}
    \label{subfig:bb}
  \end{subfigure}
  \begin{subfigure}{0.24\textwidth}
    \includegraphics[trim=315 0 315 0, clip, width=\columnwidth]{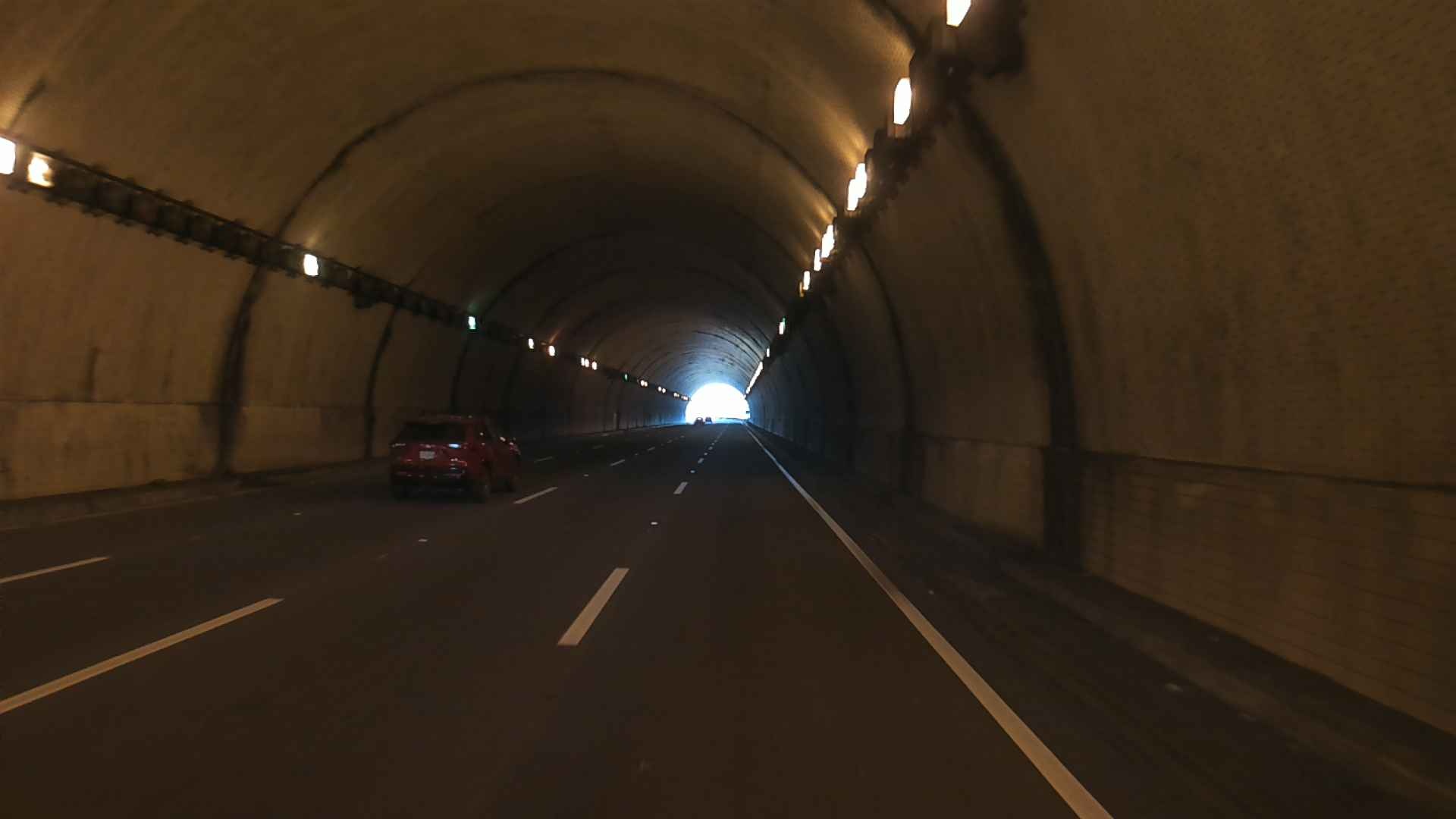}
    \caption{Robin Williams Tunnel}
    \label{subfig:rw}
  \end{subfigure}
  \begin{subfigure}{0.24\textwidth}
    \includegraphics[width=\columnwidth]{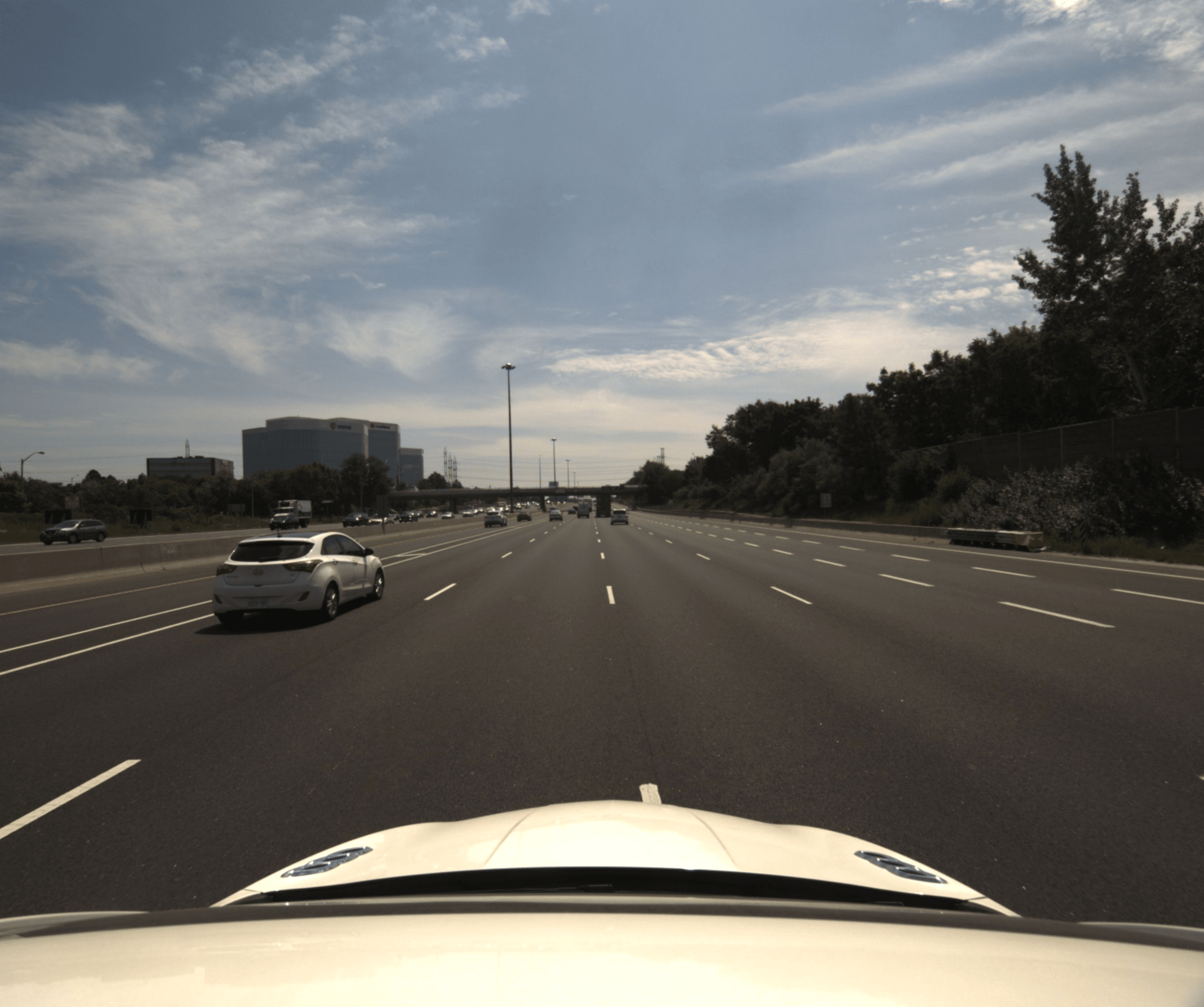}
    \caption{Ontario Highway 404}
    \label{subfig:h404}
  \end{subfigure}
  \begin{subfigure}{0.24\textwidth}
    \includegraphics[width=\columnwidth]{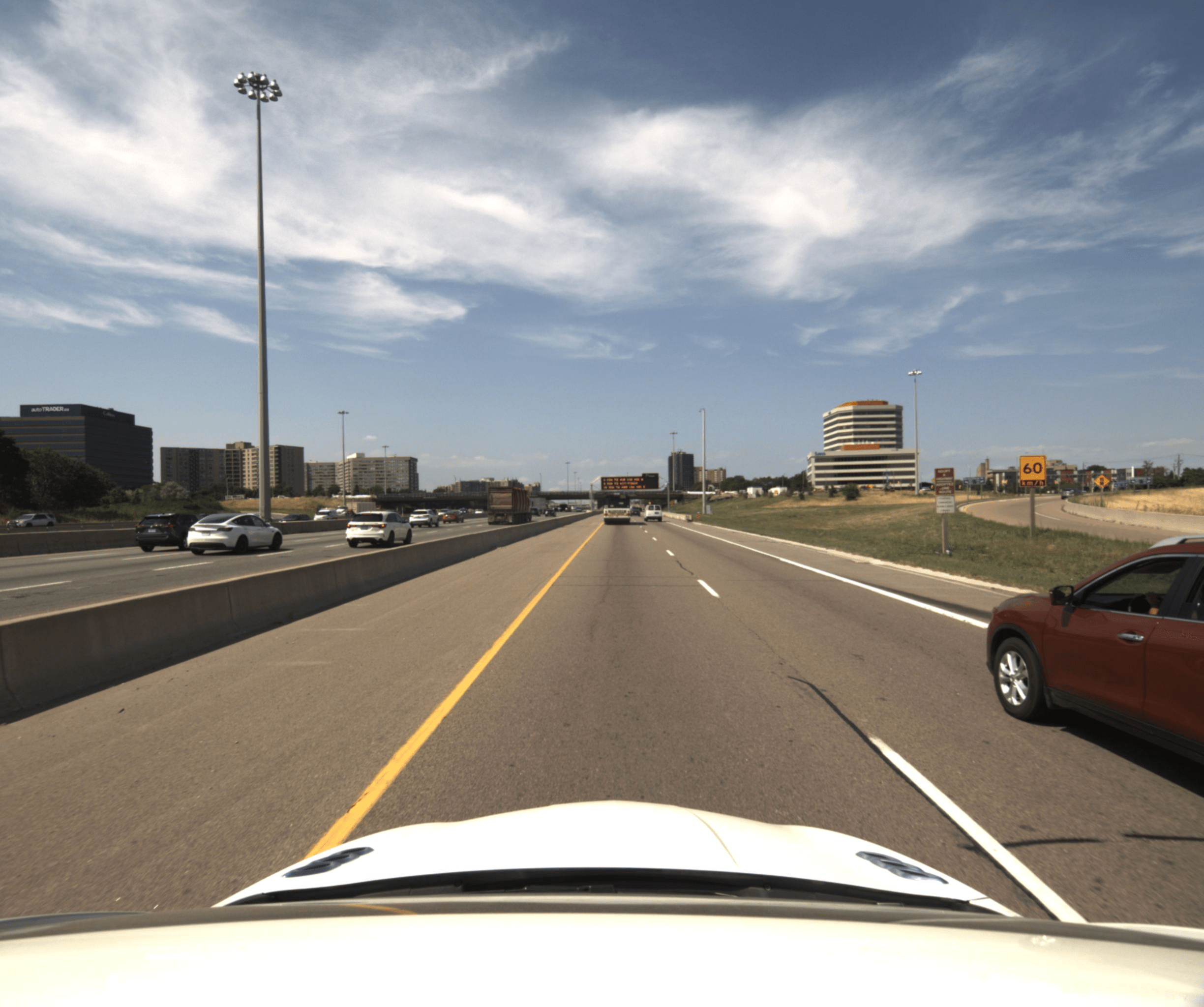}
    \caption{Ontario Highway 427}
    \label{subfig:h427}
  \end{subfigure}
  \caption{Representative scenes in the Aeva dataset. The straight tunnels (\ref{sub@subfig:bb}) and (\ref{sub@subfig:rw}) have poor geometric structure to constrain the vehicle motion in the longitudinal direction. The Ontario highway sequences have moderate geometric structure from buildings and vegetation by the side of the highway.}
  \label{fig:submaps}
\end{figure*}

\subsubsection{Doppler Velocity Factor}

The Doppler velocity factor is
\begin{eqnarray}
  \label{eq:dv_factor}
  \phi_{\text{dv}} &=& \rho \left( \sqrt{ \beta \; e^2_{\text{dv}}} \right), \\
  \label{eq:dv_error}
  e_{\text{dv}} &=& \dot{r} - \dfrac{\mbf{q}^T \mbf{D}^T}{ (\mbf{q}^T \mbf{D}^T \mbf{D} \mbf{q})^{1/2}} \mbf{D} \mbf{q}^\odot \mbs{\mathcal{T}}_{\ell v} \mbs{\varpi}^{iv}_v(t),
\end{eqnarray}
where $\mbs{\varpi}^{iv}_{v}(t)$ is the vehicle body-centric velocity queried at $t$, $\mbf{D}$ is the same projection as above, $\dot{r}$ is the Doppler velocity measurement associated with $\mbf{q}$ measured at time $t$, $\mbs{\mathcal{T}}_{\ell v} = \text{Ad}\left( \mbf{T}_{\ell v} \right)$ is the adjoint matrix of $\mbf{T}_{\ell v}$, and the $(\cdot)^\odot$ operator converts a homogeneous point to a $4\times 6$ matrix, as defined in \cite[p.246]{barfoot_se17}. $\beta=0.1$ is a constant heuristic weight and $\rho(\cdot)$ is again the Cauchy robust cost function with $c=0.05$ \cite{mactavish_crv15}. We dicard any measurement resulting in $e_{\text{dv}} \geq 2\text{m/s}$.

To derive our error term \eqref{eq:dv_error}, we first define $\dot{\mbf{q}}$ and $\dot{\mbf{q}}_i$ to be $\mbf{q}$'s velocity expressed in the lidar and inertial reference frame, respectively. Applying the \textit{transport theorem},
\begin{equation}
  \label{eq:transport_theorem}
  \dot{\mbf{q}}_i = \mbf{T}_{i\ell}(t) \left(  \dot{\mbf{q}} - {\mbs{\varpi}^{i\ell}_\ell}(t)^\wedge \mbf{q} \right).
\end{equation}
Assuming $\mbf{q}$ is static in the inertial frame such that $\dot{\mbf{q}}_i=0$, we can rearrange \eqref{eq:transport_theorem} to be (see $SE(3)$ identities in \cite{barfoot_se17})
\beqn
\dot{\mbf{q}} &=& {\mbs{\varpi}^{i\ell}_\ell}(t)^\wedge \mbf{q} \nonumber \\
&=& \mbf{q}^{\odot} \mbs{\varpi}^{i\ell}_\ell(t) \nonumber \\
\label{eq:transport_theorem_rearranged}
&=& \mbf{q}^{\odot} \mbs{\mathcal{T}}_{\ell v} \mbs{\varpi}^{iv}_v(t),
\eeqn
where \eqref{eq:transport_theorem_rearranged} relates the point velocity $\dot{\mbf{q}}$ to the body-centric velocity component of our trajectory state $\mbs{\varpi}^{iv}_v$. Next, we project $\dot{\mbf{q}}$ onto $\mbf{q}$ to obtain the predicted Doppler velocity:
\begin{equation}
  \widetilde{\dot{r}}
  = \underbrace{\dfrac{\mbf{q}^T \mbf{D}^T}{ \left( \mbf{q}^T \mbf{D}^T \mbf{D} \mbf{q} \right)^{1/2}}}_{\text{Projection } \mbf{d}}
  \mbf{D} \mbf{q}^{\odot} \mbs{\mathcal{T}}_{\ell v} \mbs{\varpi}^{iv}_v(t).
\end{equation}
Finally, taking the difference between the predicted and measured relative radial velocity gives the error function $e_{\text{dv}}$ in \eqref{eq:dv_error}. This derivation is graphically illustrated in Figure~\ref{fig:dv_derivation}.

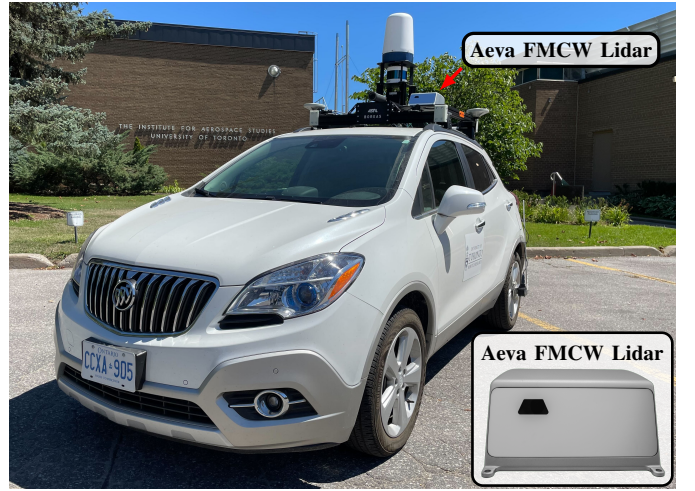
\begin{figure}[h]
  \centering
  \begin{tikzpicture} [
      arrow/.style={>=latex,red, line width=1.25pt},
      block/.style={rectangle, draw, fill=white, fill opacity=0.8, minimum width=4em, text centered, rounded corners, minimum height=1.25em, line width=1.25pt, inner sep=2.5pt}]
    \node[inner sep=0pt] (boreas) {\includegraphics[trim=300 100 0 200, clip, width=\columnwidth]{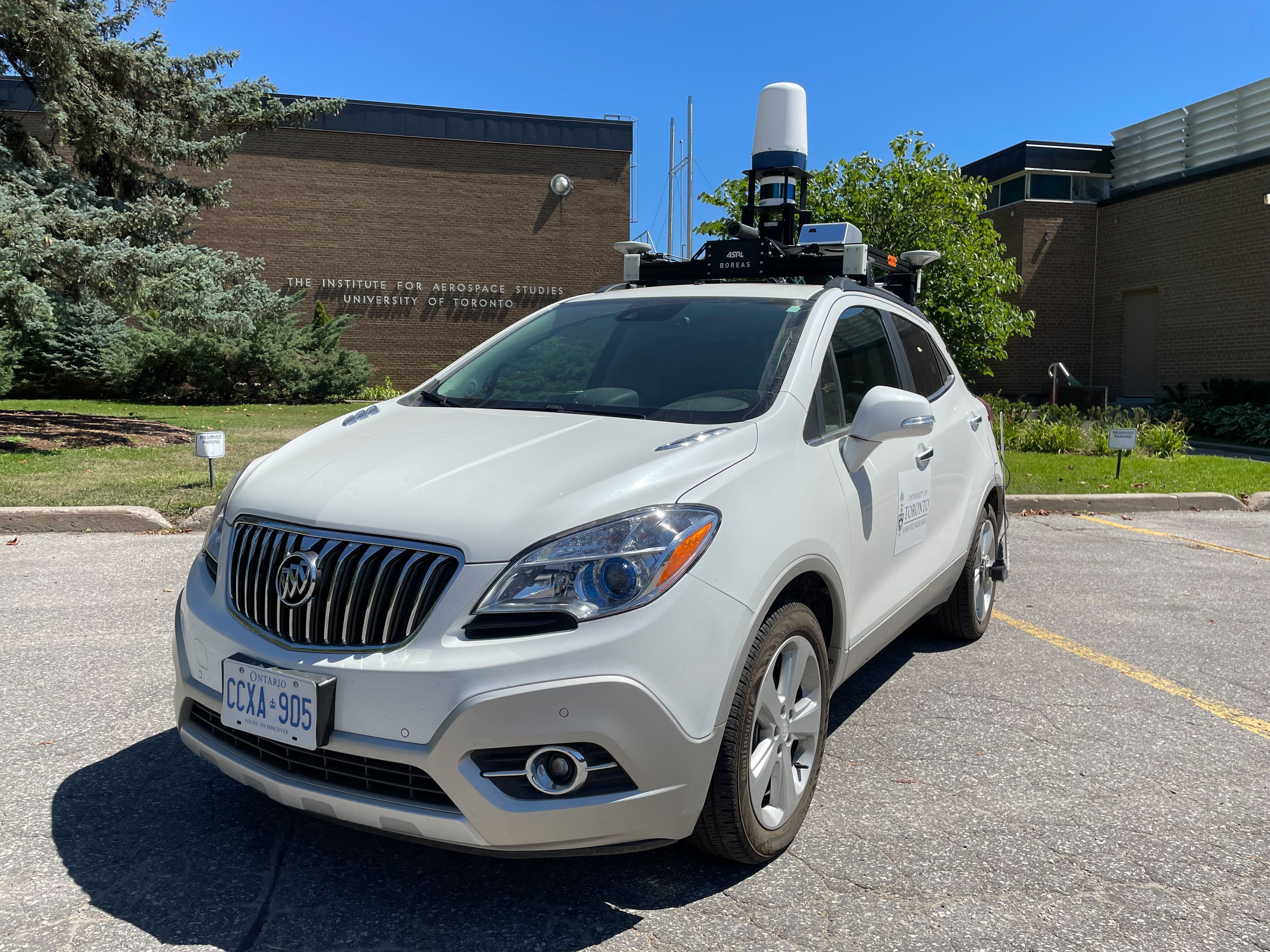}};
    \node (A) at (boreas.center) {};
    \def \L{0.75};

    \node [block, text opacity=1.0] (lidar) at ($ (A) + (29mm, 26mm) $) {\footnotesize\textbf{Aeva FMCW Lidar}};
    \draw[->, arrow] ($ (lidar.south west) + (0mm, 0mm) $) -- ($ (lidar.south west) + (-3mm, -3mm) $) {};

    \node[inner sep=0pt, draw=black, rounded corners, line width=1pt, fill=white, fill opacity=0.8, text opacity=1.0] (aeva) at ($(boreas.center) + (30.5mm, -21.8mm)$) {\includegraphics[trim=400 300 200 0, clip, width=0.3\columnwidth]{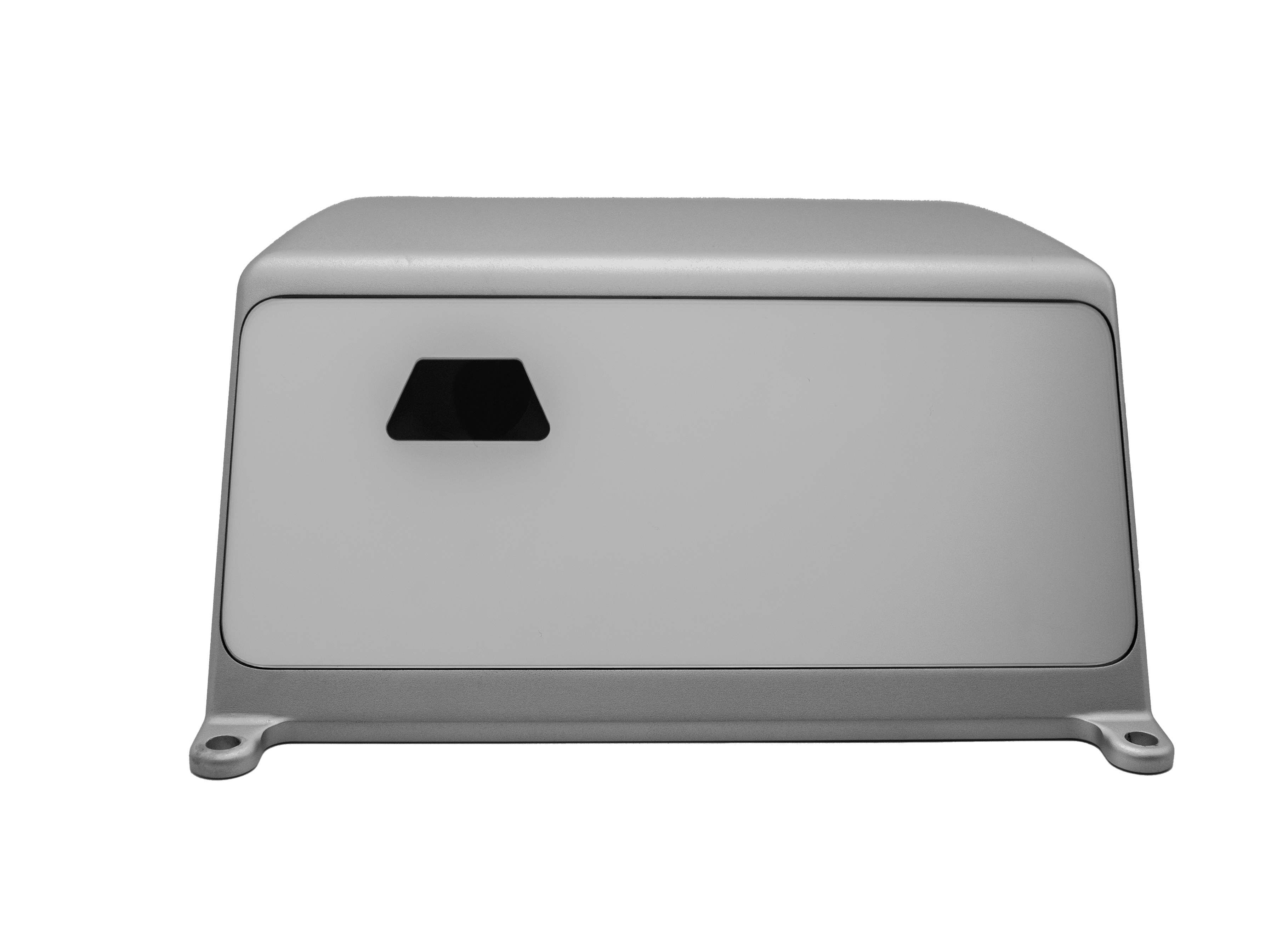}};

    \node [rectangle, text centered, inner sep=0pt] (lidar text) at ($ (aeva.center) + (0mm, 7.5mm) $) {\footnotesize\textbf{Aeva FMCW Lidar}};

  \end{tikzpicture}
  \caption{Our platform, \textit{Boreas}, is previously used to collect the \href{https://www.boreas.utias.utoronto.ca}{Boreas dataset} \cite{burnett_arxiv22} and additionally equipped with an Aeva Aeries I FMCW Lidar \cite{aeva} for use in this work.}
  \label{fig:buick}
\end{figure}

\begin{table}[h]
  \centering
  \scriptsize
  \caption{Aeva Dataset Statistics}
  \label{tab:aeva-dataset}
  \begin{tabular}{c  l r r  c }
    \toprule
    \textbf{Seq. \#} & \textbf{Location}             & \begin{tabular}[c]{@{}c@{}}\textbf{Length}\\ \textbf{[m]}\end{tabular} & \begin{tabular}[c]{@{}c@{}}\textbf{Num. of}\\ \textbf{Frames}\end{tabular} & \begin{tabular}[c]{@{}c@{}}\textbf{Geometric}\\ \textbf{Structure}\end{tabular} \\
    \midrule
    00               & Baker-Barry Tunnel (Empty)    & 860                                                                                                            & 837                                                                        & Poor                                                                            \\
    01               & Baker-Barry Tunnel (Vehicles) & 907                                                                                                            & 658                                                                        & Poor                                                                            \\
    02               & Robin Williams Tunnel         & 689                                                                                                            & 301                                                                        & Poor                                                                            \\
    03               & Brisbane Lagoon Freeway       & 4942                                                                                                           & 1762                                                                       & Poor                                                                            \\
    04               & Ontario Highway 7             & 8876                                                                                                           & 6343                                                                       & Moderate                                                                        \\
    05               & Ontario Highway 407           & 7836                                                                                                           & 4734                                                                       & Moderate                                                                        \\
    06               & Don Valley Parkway            & 10310                                                                                                          & 5083                                                                       & Moderate                                                                        \\
    07               & Ontario Highway 427           & 7238                                                                                                           & 4012                                                                       & Moderate                                                                        \\
    \bottomrule
  \end{tabular}
\end{table}

\begin{table*}[h]
  \centering
  \caption{Quantitative results on KITTI-raw/360 dataset using KITTI RTE metric. The average is computed over all segments of all sequences as in \cite{dellenbach_icra22}. Note that CT-ICP optimizes one lidar frame at a time, while our algorithm optimizes multiple frames in a sliding window. For a fair comparison, we evaluate our algorithm using the estimated poses at the front of the window (i.e., newest timestamp).}
  \label{tab:kitti-quantitative}
  \begin{tabular}{ l ? c c c c c c c c c c c ? c }
    \toprule
    \midrule
    \textbf{KITTI-raw}              & 00            & 01            & 02            & 03 (NA)       & 04            & 05            & 06            & 07            & 08            & 09            & 10            & \color{blue}{\textbf{AVG}}  \\
    \midrule
    CT-ICP \cite{dellenbach_icra22} & 0.51          & 0.81          & 0.55          & -             & 0.43          & 0.27          & \textbf{0.28} & 0.35          & \textbf{0.80} & 0.47          & \textbf{0.49} & \color{blue}{0.55}          \\
    STEAM-ICP (Ours)                & \textbf{0.49} & \textbf{0.65} & \textbf{0.50} & -             & \textbf{0.38} & \textbf{0.26} & \textbf{0.28} & \textbf{0.32} & 0.81          & \textbf{0.46} & 0.53          & \color{blue}{\textbf{0.52}} \\
    \midrule
    \textbf{KITTI-360}              & 00            & 01 (NA)       & 02            & 03            & 04            & 05            & 06            & 07            & 08 (NA)       & 09            & 10            & \color{blue}{\textbf{AVG}}  \\
    \midrule
    CT-ICP \cite{dellenbach_icra22} & \textbf{0.41} & -             & \textbf{0.38} & \textbf{0.34} & \textbf{0.65} & \textbf{0.39} & \textbf{0.42} & \textbf{0.34} & -             & 0.45          & \textbf{0.69} & \color{blue}{\textbf{0.45}} \\
    STEAM-ICP (Ours)                & \textbf{0.41} & -             & \textbf{0.38} & 0.48          & 0.69          & 0.40          & 0.43          & 0.60          & -             & \textbf{0.35} & 0.73          & \color{blue}{\textbf{0.45}} \\
    \midrule
    \bottomrule
  \end{tabular}
\end{table*}

\begin{figure*}[h]
  \centering
  \includegraphics{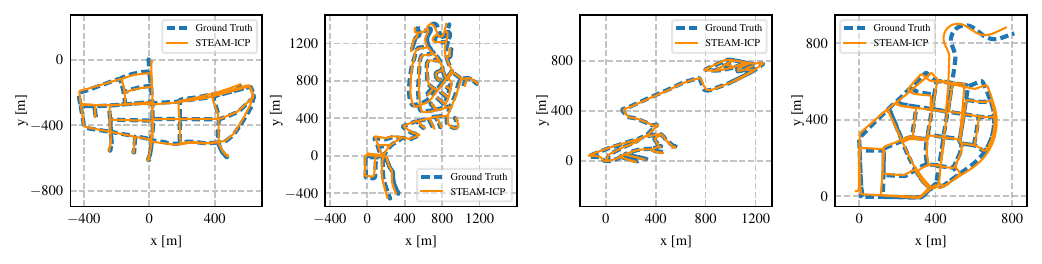}
  \vspace{-2mm}
  \caption{Trajectories estimated by STEAM-ICP from the longest four sequences (over 10000 frames) in KITTI-360 dataset: sequence 00 (11501 frames), 02 (19231 frames), 04 (11400 frames), and 09 (13955 frames).}
  \label{fig:kitti_qualitative}
\end{figure*}

At this point, we highlight the characteristics of our Doppler velocity factor. Firstly, the factor does not require the local map. Secondly, the rotational velocity of the lidar frame $\mbs{\omega}_{\ell}^{i\ell}$ is unfortunately not observable\footnote{Note that $\mbf{q}^T \mbf{D}^T \mbf{D} \mbf{q}^{\odot}$ will have its last three elements being zero, which masks the rotational component of $\mbs{\varpi}^{i\ell}_\ell$. It implies that we cannot rely solely on Doppler velocity measurements to recover the full vehicle state (pose/velocity).}. Finally, by having the vehicle's body-centric velocity as part of the state, the Doppler velocity measurement is incorporated directly into our estimator without further approximations or assumptions. While the first two characteristics have been observed by Hexsel et al. \cite{hexsel_rss22}, the last one is unique to our formulation. Hexsel et al. \cite{hexsel_rss22} derived a relation between Doppler velocity and vehicle pose using a constant-velocity approximation within the period of each frame, which is not needed in our formulation.

We treat sliding-window estimation as a factor-graph optimization problem and solve it iteratively using Gauss-Newton with Iteratively Reweighted Least Squares (IRLS) \cite{holland_cs77}. Rather than naively dropping states out of the sliding window, we explicitly marginalize out each old state. Each new state is initialized based on the GP extrapolation of our motion prior (i.e., constant velocity). The point-to-plane association of keypoints in the latest frame and the plane normals are re-computed after every five Gauss-Newton iterations.


\section{Experiments}

\subsection{Datasets}
We demonstrate our method and make comparisons to existing work on several real-world data sequences, which we categorize into two datasets: KITTI-raw/360 and Aeva. All data sequences provide accurate pose estimates from a GNSS-INS system for evaluation\footnote{For ground truth pose generation, we refer the readers to \cite{geiger_cvpr12} for KITTI-raw/360, \cite{hexsel_rss22} for Aeva sequence 00-03, and \cite{burnett_arxiv22} for Aeva sequence 04-07.}.

\subsubsection{KITTI-raw/360 dataset}

contains 19 sequences of raw lidar frames from the publicly available KITTI dataset \cite{geiger_cvpr12} and its successor KITTI-360 \cite{xie_cvpr16}. The lidar used in this dataset is a Velodyne HDL-64, which is not an FMCW lidar. The sequences in this dataset were collected from urban/suburban environments and are rich in geometric structure for lidar motion estimation. We apply our estimator to the variation of the dataset made available by Dellenbach et al. \cite{dellenbach_icra22}.

\subsubsection{Aeva dataset}

contains 8 sequences of lidar frames from an Aeva Aeries I FMCW Lidar that produces Doppler velocity measurements\footnote{We have published these sequences, which can be found at \url{https://github.com/utiasASRL/steam_icp}.}. Aeva Aeries I FMCW Lidar has a horizontal field-of-view of $120^\circ$, a vertical field-of-view of $30^\circ$, a 300m maximum operating range, a Doppler velocity measurement precision of 3cm/s, and a sampling rate of 10Hz.

Sequences 00-03 are relatively short sequences purposely collected in environments with poor geometric structure (e.g., tunnels, freeways), which have been used in previous work \cite{hexsel_rss22}, but required an additional IMU to correct for the scanning-while-moving nature of the sensor. Sequences 04-07 are longer sequences collected using our data collection platform, \textit{Boreas} (Figure~\ref{fig:buick}), on Toronto, Ontario highways with moderate geometric structure. Table~\ref{tab:aeva-dataset} has a summary of the dataset statistics, and Figure~\ref{fig:submaps} shows some representative scenes.

\begin{table*}[h]
  \centering
  \caption{Quantitative results on Aeva dataset using both KITTI RTE and Frame-to-Frame RTE metrics. Note that all algorithms (including Doppler-ICP) are evaluated using motion-distorted frames. We exclude the first 60 frames of each sequence from evaluation because the vehicle does not start at zero velocity in these sequences, which causes the local map to be initialized from motion-distorted frames. Results of \textbf{Seq. 04-07 (Range-Limited)} were obtained by limiting the range of the lidar frames to 40m.}
  \label{tab:aeva-quantitative}
  \begin{tabular}{ l ? c c c c c ? c c c c c }
    \toprule
    \midrule
                                        & \multicolumn{5}{c|}{KITTI RTE [\%]} & \multicolumn{5}{c}{Frame-to-Frame RTE [m]}                                                                                                                                                                       \\
    \midrule
    \textbf{Sequences 00-03}            & 00                                  & 01                                         & 02            & 03            & {\color{blue}\textbf{AVG}}  & 00              & 01              & 02              & 03              & {\color{blue}\textbf{AVG}}    \\
    \midrule
    Doppler-ICP \cite{hexsel_rss22}     & \textbf{1.66}                       & \textbf{2.60}                              & 1.03          & 1.72          & {\color{blue}\textbf{1.80}} & 0.0246          & 0.0254          & 0.0380          & 0.0494          & {\color{blue}0.0402}          \\
    CT-ICP \cite{dellenbach_icra22}     & 2.83                                & 12.26                                      & 9.11          & \textbf{1.54} & {\color{blue}3.35}          & 0.0401          & 0.3753          & 0.2446          & 0.0801          & {\color{blue}0.1827}          \\
    STEAM-ICP (Ours)                    & 2.28                                & 12.86                                      & 22.74         & 2.10          & {\color{blue}4.16}          & 0.0541          & 0.4134          & 0.6076          & 0.2892          & {\color{blue}0.3180}          \\
    STEAM-DICP (Ours)                   & 2.35                                & \textbf{2.60}                              & \textbf{0.74} & 1.70          & {\color{blue}1.88}          & \textbf{0.0180} & \textbf{0.0211} & \textbf{0.0299} & \textbf{0.0362} & {\color{blue}\textbf{0.0299}} \\
    \midrule

    \textbf{Sequences 04-07}            & 04                                  & 05                                         & 06            & 07            & {\color{blue}\textbf{AVG}}  & 04              & 05              & 06              & 07              & {\color{blue}\textbf{AVG}}    \\
    \midrule
    Doppler-ICP \cite{hexsel_rss22}     & 3.86                                & 2.81                                       & 2.13          & 2.93          & 3.00                        & 0.0864          & 0.0485          & 0.0140          & 0.2355          & {\color{blue}0.1181}          \\
    CT-ICP \cite{dellenbach_icra22}     & 0.34                                & 0.34                                       & 0.41          & 0.48          & 0.38                        & 0.0194          & 0.0202          & 0.0198          & 0.0216          & {\color{blue}0.0202}          \\
    STEAM-ICP (Ours)                    & 0.38                                & \textbf{0.29}                              & 0.36          & \textbf{0.36} & \textbf{0.35}               & 0.0211          & 0.0250          & 0.0230          & 0.0298          & {\color{blue}0.0244}          \\
    STEAM-DICP (Ours)                   & \textbf{0.33}                       & 0.46                                       & \textbf{0.30} & 0.37          & 0.36                        & \textbf{0.0064} & \textbf{0.0119} & \textbf{0.0081} & \textbf{0.0201} & {\color{blue}\textbf{0.0119}} \\

    \midrule

    \textbf{Seq. 04-07 (Range-Limited)} & 04                                  & 05                                         & 06            & 07            & {\color{blue}\textbf{AVG}}  & 04              & 05              & 06              & 07              & {\color{blue}\textbf{AVG}}    \\
    \midrule
    Doppler-ICP \cite{hexsel_rss22}     & 13.73                               & 6.29                                       & 2.72          & 6.95          & {\color{blue}7.96}          & 0.2555          & 0.1679          & \textbf{0.0198} & 0.4047          & {\color{blue}0.2444}          \\
    CT-ICP \cite{dellenbach_icra22}     & 10.90                               & 56.44                                      & 66.64         & 5.39          & {\color{blue}33.87}         & 0.0857          & 1.9122          & 1.8926          & 0.0637          & {\color{blue}1.3279}          \\
    STEAM-ICP (Ours)                    & 67.48                               & 3.48                                       & 2.57          & 3.99          & {\color{blue}24.50}         & 1.6030          & 0.1107          & 0.0644          & 0.1778          & {\color{blue}0.9056}          \\
    STEAM-DICP (Ours)                   & \textbf{2.25}                       & \textbf{3.11}                              & \textbf{2.28} & \textbf{2.27} & {\color{blue}\textbf{2.44}} & \textbf{0.0173} & \textbf{0.0337} & 0.0211          & \textbf{0.0453} & {\color{blue}\textbf{0.0297}} \\
    \midrule
    \bottomrule
  \end{tabular}
\end{table*}

\begin{figure*}[h]
  \centering
  \begin{tikzpicture}[
      boundary/.style={>=latex,blue, line width=1.25pt},
      custom arrow/.style={single arrow, minimum height=12mm, minimum width=1mm, single arrow head extend=1.5mm, scale=0.15, fill=blue},
      block/.style={rectangle, blue, minimum width=4mm, align=center, inner sep=0pt},
      rotate border/.style={shape border uses incircle, shape border rotate=#1},
    ]

    \node[inner sep=0pt] (p1) at (0mm, 0mm) {\includegraphics[width=0.45\textwidth]{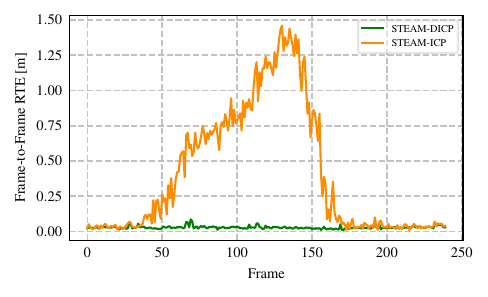}};
    \draw[dashed, boundary] ($ (p1) + (-15mm, -15mm)$) -- ($ (p1) + (-15mm, 25mm)$) {};
    \draw[dashed, boundary] ($ (p1) + (19mm, -15mm)$) -- ($ (p1) + (19mm, 25mm)$) {};
    \node[block] (p1ts) at ($ (p1) + (-15mm, 26mm)$) {\small Tunnel Start};
    \node[block] (p1te) at ($ (p1) + (19mm, 26mm)$) {\small Tunnel End};

    \node[inner sep=0pt] (p2) at ($ (p1) + (65mm, 0mm)$) {\includegraphics[width=0.282\textwidth]{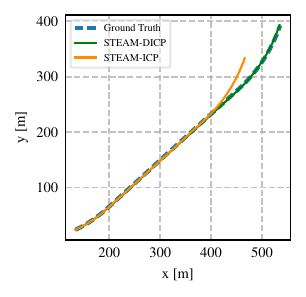}};
    \node[custom arrow, rotate border=130] (p2ts) at ($ (p2) + (-5mm, -10mm)$) {};
    \node[block] (p2tst) at ($ (p2ts) + (4mm, -3mm)$) {\footnotesize Tunnel Start};

    \node[custom arrow, rotate border=130] (p2te) at ($ (p2) + (14mm, 7mm)$) {};
    \node[block] (p2tet) at ($ (p2te) + (2mm, -3mm)$) {\footnotesize Tunnel End};

    \node[inner sep=0pt] (p3) at ($ (p2) + (50mm, 3.5mm)$) {\includegraphics[width=0.43\columnwidth]{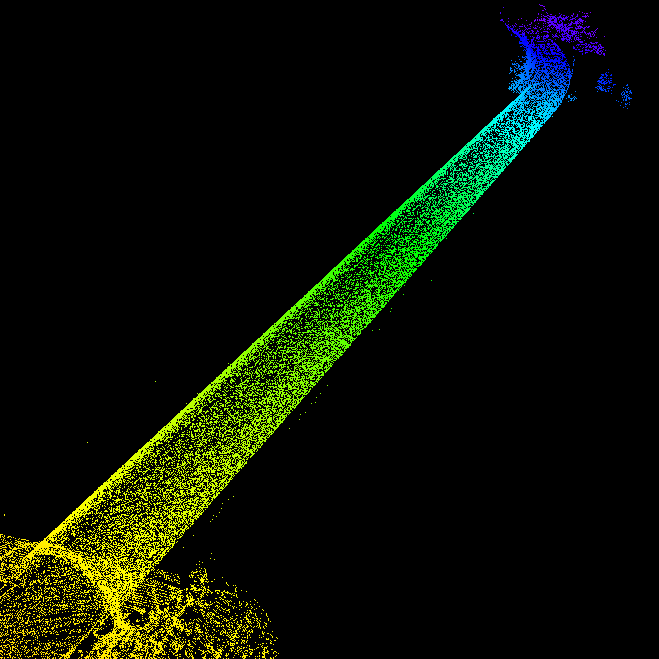}};
    \node[block, white, rotate=0] (p3t) at ($ (p3) + (8mm, -15mm)$) {\small \textbf{300m Tunnel}};

  \end{tikzpicture}
  \vspace{-2mm}
  \caption{This figure shows a comparison of Frame-to-Frame RTE between STEAM-ICP and STEAM-DICP on the Robin Williams Tunnel sequence (\textit{left}), the corresponding trajectories versus ground truth (\textit{middle}), and the point cloud map built from STEAM-DICP (\textit{right}). Due to the poor geometric structure in the longitudinal direction, STEAM-ICP under-estimates the length of the trajectory inside the tunnel, while STEAM-DICP remains unaffected thanks to the correction from Doppler velocity measurements.}
  \label{fig:seq02_local_rte}
\end{figure*}
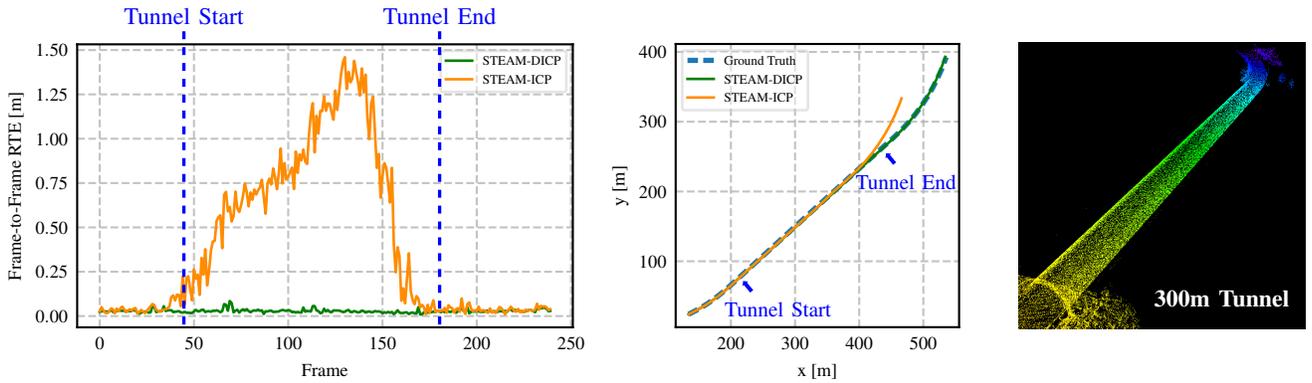

\begin{figure*}[h]
  \centering
  \includegraphics{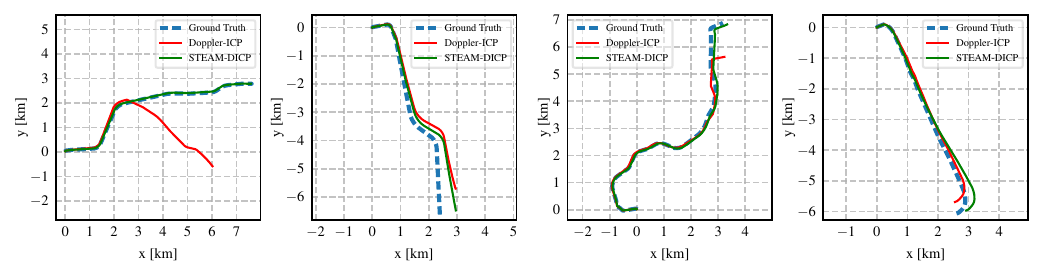}
  \vspace{-2mm}
  \caption{Trajectories estimated by STEAM-DICP and Doppler-ICP from sequences 04-07 in the Aeva dataset without limiting the range of the sensor. Doppler-ICP exhibits higher drifts due to using a frame-to-frame registration approach and not accounting for motion distortion.}
  \label{fig:aeva_qualitative}
\end{figure*}

\subsection{Evaluation}

We use the KITTI Relative Translation Error (KITTI RTE), which averages the translation error over path segments of lengths 100m to 800m in 100m intervals. For the Aeva dataset, we additionally include evaluation using the Frame-to-Frame Relative Translation Error (Frame-to-Frame RTE) metric \cite{prokhorov_mva19}, which was used in \cite{hexsel_rss22}. In the interest of space, we omit showing the rotation error in the main text since the translation error is where our comparisons differ the most. Interested readers can find our full evaluation results in our supplementary material \cite{wu_arxiv22}.

We name the two variants of our algorithm using and not using the Doppler velocity measurements STEAM-DICP and STEAM-ICP, respectively, after the continuous-time trajectory estimation framework, STEAM, presented in \cite{anderson_iros15}.

On the KITTI-raw/360 dataset, we compare STEAM-ICP with CT-ICP \cite{dellenbach_icra22}. CT-ICP, to the best of our knowledge, is the current state-of-the-art continuous-time lidar-only odometry algorithm. It achieves similar performance on the KITTI odometry benchmark as other state-of-the-art methods (e.g., LOAM \cite{zhang_ar17}, IMLS-SLAM \cite{deschaud_icra18}, and MULLS \cite{pan_icra21}) using lidar frames without motion distortion and demonstrates better performance using lidar frames with motion distortion. On the Aeva dataset, we compare STEAM-DICP with STEAM-ICP, CT-ICP, and Doppler-ICP \cite{hexsel_rss22}. Note that Doppler-ICP is the only existing algorithm that also uses Doppler velocity measurements to aid lidar odometry.

\subsection{Results}

Table~\ref{tab:kitti-quantitative} shows quantitative results on the KITTI-raw/360 dataset. Compared to CT-ICP, our odometry achieves comparable performance, demonstrating that we are performing at state of the art using non-FMCW lidars (i.e., no Doppler velocity measurements). Example plots of the estimated trajectory are shown in Figure~\ref{fig:kitti_qualitative}.

Quantitative results on the Aeva dataset are shown in Table~\ref{tab:aeva-quantitative}. On the difficult sequences with insufficient geometric structure (i.e., sequences 00-03), both CT-ICP and STEAM-ICP (i.e., methods that do not use Doppler velocity measurements) perform poorly. Figure~\ref{fig:seq02_local_rte} highlights the problem in sequence 02 (Robin Williams Tunnel), where we see that the Frame-to-Frame RTE for both CT-ICP and STEAM-ICP increases dramatically during the tunnel portion of the sequence.

On sequences with moderate geometric structure (i.e., sequences 04-07), methods that do not use the velocity measurements perform well. Doppler-ICP performs worse overall due to using a frame-to-frame approach (i.e., no accumulated local maps) and not accounting for the motion in each lidar frame. STEAM-DICP performs similarly to STEAM-ICP and CT-ICP, showing that the velocity measurements are not as impactful on sequences with sufficient geometry. However, we still see minor improvements in Frame-to-Frame RTE, suggesting that the velocity measurements overall help improve performance.

Another way to highlight the benefit of the Doppler velocity measurements is to limit the range of the lidar artificially. The majority of the distinctive structure in the environment is off-road and further away from the sensor, therefore limiting the range increases the difficulty for estimation. We see in the Range-Limited section of Table~\ref{tab:aeva-quantitative} that methods using the Doppler velocity measurements, Doppler-ICP and STEAM-DICP, are the most robust to this artificial increase in difficulty, with ours (STEAM-DICP) performing the best.

\subsection{Implementation}

We implemented our back-end (see Section~\ref{sec:trajectory-estimation}) for continuous-time trajectory estimation using an open-source C++ library, \href{https://github.com/utiasASRL/steam}{STEAM}\footnote{STEAM library: \url{https://github.com/utiasASRL/steam}} \cite{anderson_iros15}. The front-end that processes the lidar data was made to be the same as CT-ICP by using the C++ library that the authors have made publicly available\footnote{CT-ICP library: \url{https://github.com/jedeschaud/ct_icp}}. We use the same parameters as CT-ICP for keypoint extraction, neighborhood search, normal estimation, and map building to maintain a fair comparison. Parameters of the trajectory estimation back-end were determined empirically.



Our implementation is currently not quite real-time capable, running at approximately 5Hz on the KITTI sequences and approximately 2Hz on the Aeva sequences after incorporating the Doppler velocity measurements\footnote{We run experiments using an Intel Xeon E5-2698 v4 2.2 GHz processor with 20 physical cores. We parallelize trajectory interpolation and Jacobian computation using 20 threads.}. The current bottleneck is the processing of each motion prior and measurement factor in STEAM, which was designed with more focus on generalization rather than computational speed. We strongly believe that a real-time capable implementation can be achieved by implementing an estimator specific to our problem, e.g., by hardcoding Jacobian computations instead of relying on STEAM's built-in automatic differentiation.





\section{Conclusions}

In this letter, we presented a continuous-time lidar odometry algorithm leveraging the Doppler velocity measurements from an FMCW lidar to aid odometry. Our algorithm combines the lidar data processing front-end in \cite{dellenbach_icra22} with our STEAM continuous-time estimator as the back-end, efficiently estimating the vehicle trajectory using \ac{GP} regression. Through continuous-time estimation, our algorithm handles the motion distortion problem of mechanically actuated lidars. Incorporating the Doppler velocity information helps prevent odometry failures under geometrically degenerate conditions. By estimating vehicle pose and body-centric velocity, Doppler velocity measurements are incorporated directly into the estimator without further approximations or assumptions. Using both publicly available and our own datasets, we demonstrated state-of-the-art lidar odometry performance under nominal conditions with and without velocity information. On kilometer-scale highway sequences, we demonstrated superior performance over the only existing lidar odometry method that also uses Doppler velocity information under nominal and geometrically degenerate conditions.

There are multiple directions for future work to improve the proposed algorithm further. Currently, we apply robust cost functions to the point-to-plane error $e_\text{p2p}$ and the Doppler velocity error $e_\text{dv}$ individually for outlier rejection. Alternatively, one can apply a single robust cost function to $ \begin{bmatrix} e_{\text{p2p}} & e_{\text{dv}} \end{bmatrix}^T$ to reject outliers using both sources of information (in the same spirit as the \textit{Dynamic Point Outlier Rejection} scheme in \cite{hexsel_rss22}). The velocity measurements can also be used by the data processing front-end to segment and remove points from moving objects before data association (e.g., by using the method of \cite{guo_cacre22}). In addition, one can replace the \ac{WNOA} GP prior used in this work with the WNOJ prior \cite{tang_ral19} or the Singer prior \cite{wong_ral20}, which have shown to improve lidar odometry in nominal conditions.

\bibliography{IEEEabrv,references}

\clearpage
\onecolumn

\section*{Supplementary Material}
\label{sec:supplementary material}

This supplementary material presents the full quantitative results on the Aeva dataset using both KITTI Relative Pose Error metric (Table~\ref{tab:aeva-kitti-3d}) and Frame-to-Frame Relative Pose Error metric (Table~\ref{tab:aeva-rpe-3d}) \cite{prokhorov_mva19}. Note again that all algorithms are evaluated using motion-distorted frames, and we exclude the first 60 frames of each sequence from evaluation. Results of \textbf{Seq. 04-07 (Range-Limited)} were obtained by limiting the range of the lidar frames to 40m.

Results of the translation error have been reported and discussed in the main text. Regarding the rotation error, we see that Doppler-ICP seems to do slightly better in rotation on the Frame-to-Frame metric. However, the overall difference between Doppler-ICP and STEAM-DICP is not significant, and our algorithm outperforms theirs on the KITTI metric, which averages rotation error over longer trajectory segments.

\begin{table}[hp]
  \centering
  \caption{Quantitative results on Aeva dataset using KITTI Relative Pose Error metric.}
  \label{tab:aeva-kitti-3d}
  \begin{tabular}{ l ? c c c c c ? c c c c c }
    \toprule
    \midrule
                                        & \multicolumn{5}{c|}{Translation [\%]} & \multicolumn{5}{c}{Rotation [deg/m]}                                                                                                                                                                       \\
    \midrule
    \textbf{Sequences 00-03}            & 00                                    & 01                                   & 02            & 03            & {\color{blue}\textbf{AVG}}  & 00              & 01              & 02              & 03              & {\color{blue}\textbf{AVG}}    \\
    \midrule
    Doppler-ICP \cite{hexsel_rss22}     & \textbf{1.66}                         & \textbf{2.60}                        & 1.03          & 1.72          & {\color{blue}\textbf{1.80}} & 0.0330          & \textbf{0.0143} & 0.0335          & 0.0064          & {\color{blue}0.0122}          \\
    CT-ICP \cite{dellenbach_icra22}     & 2.83                                  & 12.26                                & 9.11          & \textbf{1.54} & {\color{blue}3.35}          & 0.0085          & 0.0148          & \textbf{0.0121} & \textbf{0.0038} & {\color{blue}\textbf{0.0062}} \\
    STEAM-ICP (Ours)                    & 2.28                                  & 12.86                                & 22.74         & 2.10          & {\color{blue}4.16}          & 0.0078          & 0.0155          & 0.0124          & 0.0040          & {\color{blue}0.0063}          \\
    STEAM-DICP (Ours)                   & 2.35                                  & \textbf{2.60}                        & \textbf{0.74} & 1.70          & {\color{blue}1.88}          & \textbf{0.0077} & 0.0166          & 0.0137          & 0.0040          & {\color{blue}0.0065}          \\
    \midrule

    \textbf{Sequences 04-07}            & 04                                    & 05                                   & 06            & 07            & {\color{blue}\textbf{AVG}}  & 04              & 05              & 06              & 07              & {\color{blue}\textbf{AVG}}    \\
    \midrule
    Doppler-ICP \cite{hexsel_rss22}     & 3.86                                  & 2.81                                 & 2.13          & 2.93          & {\color{blue}3.00}          & 0.0115          & 0.0104          & 0.0079          & 0.0060          & {\color{blue}0.0093}          \\
    CT-ICP \cite{dellenbach_icra22}     & 0.34                                  & 0.34                                 & 0.41          & 0.48          & {\color{blue}0.38}          & \textbf{0.0010} & 0.0012          & 0.0014          & 0.0015          & {\color{blue}0.0012}          \\
    STEAM-ICP (Ours)                    & 0.38                                  & \textbf{0.29}                        & 0.36          & \textbf{0.36} & {\color{blue}\textbf{0.35}} & 0.0012          & \textbf{0.0009} & 0.0011          & \textbf{0.0012} & {\color{blue}\textbf{0.0011}} \\
    STEAM-DICP (Ours)                   & \textbf{0.33}                         & 0.46                                 & \textbf{0.30} & 0.37          & {\color{blue}0.36}          & \textbf{0.0010} & 0.0013          & \textbf{0.0009} & \textbf{0.0012} & {\color{blue}\textbf{0.0011}} \\
    \midrule

    \textbf{Seq. 04-07 (Range-Limited)} & 04                                    & 05                                   & 06            & 07            & {\color{blue}\textbf{AVG}}  & 04              & 05              & 06              & 07              & {\color{blue}\textbf{AVG}}    \\
    \midrule
    Doppler-ICP \cite{hexsel_rss22}     & 13.73                                 & 6.29                                 & 2.72          & 6.95          & {\color{blue}7.96}          & 0.0350          & 0.0224          & 0.0103          & 0.0113          & {\color{blue}0.0213}          \\
    CT-ICP \cite{dellenbach_icra22}     & 10.90                                 & 56.44                                & 66.64         & 5.39          & {\color{blue}33.87}         & 0.0315          & 0.2084          & 0.2239          & 0.0165          & {\color{blue}0.1158}          \\
    STEAM-ICP (Ours)                    & 67.48                                 & 3.48                                 & 2.57          & 3.99          & {\color{blue}24.50}         & 0.1218          & \textbf{0.0087} & 0.0072          & 0.0076          & {\color{blue}0.0455}          \\
    STEAM-DICP (Ours)                   & \textbf{2.25}                         & \textbf{3.11}                        & \textbf{2.28} & \textbf{2.27} & {\color{blue}\textbf{2.44}} & \textbf{0.0062} & 0.0090          & \textbf{0.0069} & \textbf{0.0071} & {\color{blue}\textbf{0.0072}} \\
    \midrule
    \bottomrule
  \end{tabular}
\end{table}

\begin{table}[hp]
  \centering
  \caption{Quantitative results on Aeva dataset using Frame-to-Frame Relative Pose Error metric.}
  \label{tab:aeva-rpe-3d}
  \begin{tabular}{ l ? c c c c c ? c c c c c }
    \toprule
    \midrule
                                        & \multicolumn{5}{c|}{Translation [m]} & \multicolumn{5}{c}{Rotation [deg]}                                                                                                                                                                             \\
    \midrule
    \textbf{Sequences 00-03}            & 00                                   & 01                                 & 02              & 03              & {\color{blue}\textbf{AVG}}    & 00              & 01              & 02              & 03              & {\color{blue}\textbf{AVG}}    \\
    \midrule
    Doppler-ICP \cite{hexsel_rss22}     & 0.0246                               & 0.0254                             & 0.0380          & 0.0494          & {\color{blue}0.0402}          & 0.1357          & \textbf{0.1670} & 0.1655          & \textbf{0.0827} & {\color{blue}\textbf{0.1163}} \\
    CT-ICP \cite{dellenbach_icra22}     & 0.0401                               & 0.3753                             & 0.2446          & 0.0801          & {\color{blue}0.1827}          & 0.1907          & 0.2865          & 0.1593          & 0.1163          & {\color{blue}0.1675}          \\
    STEAM-ICP (Ours)                    & 0.0541                               & 0.4134                             & 0.6076          & 0.2892          & {\color{blue}0.3180}          & \textbf{0.1322} & 0.1855          & 0.1503          & 0.1195          & {\color{blue}0.1366}          \\
    STEAM-DICP (Ours)                   & \textbf{0.0180}                      & \textbf{0.0211}                    & \textbf{0.0299} & \textbf{0.0362} & {\color{blue}\textbf{0.0299}} & 0.1384          & 0.1821          & \textbf{0.1475} & 0.1125          & {\color{blue}0.1336}          \\
    \midrule

    \textbf{Sequences 04-07}            & 04                                   & 05                                 & 06              & 07              & {\color{blue}\textbf{AVG}}    & 04              & 05              & 06              & 07              & {\color{blue}\textbf{AVG}}    \\
    \midrule
    Doppler-ICP \cite{hexsel_rss22}     & 0.0864                               & 0.0485                             & 0.0140          & 0.2355          & {\color{blue}0.1181}          & 0.0557          & 0.0453          & 0.0655          & 0.0559          & {\color{blue}0.0558}          \\
    CT-ICP \cite{dellenbach_icra22}     & 0.0194                               & 0.0202                             & 0.0198          & 0.0216          & {\color{blue}0.0202}          & 0.0460          & 0.0401          & \textbf{0.0582} & 0.0501          & {\color{blue}0.0485}          \\
    STEAM-ICP (Ours)                    & 0.0211                               & 0.0250                             & 0.0230          & 0.0298          & {\color{blue}0.0244}          & 0.0460          & 0.0410          & 0.0603          & 0.0489          & {\color{blue}0.0490}          \\
    STEAM-DICP (Ours)                   & \textbf{0.0064}                      & \textbf{0.0119}                    & \textbf{0.0081} & \textbf{0.0201} & {\color{blue}\textbf{0.0119}} & \textbf{0.0447} & \textbf{0.0391} & 0.0587          & \textbf{0.0474} & {\color{blue}\textbf{0.0474}} \\
    \midrule

    \textbf{Seq. 04-07 (Range-Limited)} & 04                                   & 05                                 & 06              & 07              & {\color{blue}\textbf{AVG}}    & 04              & 05              & 06              & 07              & {\color{blue}\textbf{AVG}}    \\
    \midrule
    Doppler-ICP \cite{hexsel_rss22}     & 0.2555                               & 0.1679                             & \textbf{0.0198} & 0.4047          & {\color{blue}0.2444}          & \textbf{0.0917} & \textbf{0.0673} & \textbf{0.0688} & \textbf{0.0609} & {\color{blue}\textbf{0.0741}} \\
    CT-ICP \cite{dellenbach_icra22}     & 0.0857                               & 1.9122                             & 1.8926          & 0.0637          & {\color{blue}1.3279}          & 0.1300          & 71.1626         & 77.2449         & 0.1086          & {\color{blue}36.2160}         \\
    STEAM-ICP (Ours)                    & 1.6030                               & 0.1107                             & 0.0644          & 0.1778          & {\color{blue}0.9056}          & 0.8207          & 0.1058          & 0.1211          & 0.0944          & {\color{blue}0.3328}          \\
    STEAM-DICP (Ours)                   & \textbf{0.0173}                      & \textbf{0.0337}                    & 0.0211          & \textbf{0.0453} & {\color{blue}\textbf{0.0297}} & 0.0923          & 0.0844          & 0.1059          & 0.0852          & {\color{blue}0.0925}          \\
    \midrule
    \bottomrule
  \end{tabular}
\end{table}

\end{document}